\documentclass[conference]{IEEEtran}
\IEEEoverridecommandlockouts
\usepackage{cite}
\usepackage{xcolor}
\usepackage{amsmath,amssymb,amsfonts}
\usepackage{diagbox}
\usepackage{algorithmic}
\usepackage{graphicx}
\usepackage{textcomp}

\usepackage{float} 
\usepackage{xcolor}
\usepackage{diagbox}
\usepackage{url}
\newtheorem{defn}{Definition}
\usepackage{subfigure}
\usepackage{booktabs}
\usepackage{threeparttable}
 \usepackage{multirow}
\def\BibTeX{{\rm B\kern-.05em{\sc i\kern-.025em b}\kern-.08em
    T\kern-.1667em\lower.7ex\hbox{E}\kern-.125emX}}

\begin{document}

\title{MMKGR: Multi-hop Multi-modal\\ Knowledge Graph Reasoning\\
}

\author
{
	Shangfei Zheng{\small$^\dag$}\hspace*{10pt}
	Weiqing Wang{\small$^\dag\dag$} \hspace*{10pt}
	Jianfeng Qu{\small$^\dag$}\hspace*{10pt}
	Hongzhi Yin{\small$^\ddag$}\hspace*{10pt}
	Wei Chen{\small$^\dag$}{$^*$}\hspace*{10pt}\thanks{* These authors are corresponding authors.} Lei Zhao{\small$^\dag $}{$^*$}\hspace*{10pt}\hspace*{10pt}\\
	\fontsize{10}{10}\selectfont\itshape $~^\dag$School of Computer Science and Technology, Soochow University\\
	\fontsize{10}{10}\selectfont\itshape
	
	$~^\dag\dag$ Department of Data Science and AI, Monash University\\
		\fontsize{10}{10}\selectfont\itshape
	$~^\ddag$ School of Information Technology and Electrical Engineering, The University of Queensland\\

	\fontsize{9}{9}\selectfont\ttfamily\upshape$~^\dag$sfzhengsuda@stu.suda.edu.cn\hspace*{10pt}$~^\dag\dag$Teresa.Wang@monash.edu\\
	$~^\ddag$db.hongzhi@gmail.com\hspace*{10pt}$~^\dag$\{jfqu, robertchen, zhaol\}@suda.edu.cn
}

\maketitle

\begin{abstract}
Multi-modal knowledge graphs (MKGs) include not only the relation triplets, but also related multi-modal auxiliary data (i.e., texts and images), which enhance the diversity of knowledge. However, the natural incompleteness has significantly hindered the applications of MKGs. To tackle the problem, existing studies employ the embedding-based reasoning models to infer the missing knowledge after fusing the multi-modal features. However, the reasoning performance of these  methods is limited due to the following problems: (1) ineffective fusion of multi-modal auxiliary features; (2) lack of complex reasoning ability as well as inability to conduct the multi-hop reasoning which is able to infer more missing knowledge. To overcome these problems, we propose a novel model entitled MMKGR (\textbf{M}ulti-hop \textbf{M}ulti-modal \textbf{K}nowledge \textbf{G}raph \textbf{R}easoning). Specifically, the model contains the following two components: (1) a unified gate-attention network which is designed to generate effective multi-modal complementary features through sufficient attention interaction and noise reduction; (2) a complementary feature-aware reinforcement learning method which is proposed to predict missing elements by performing the multi-hop reasoning process, based on the features obtained in component (1). The experimental results demonstrate that MMKGR  outperforms the state-of-the-art approaches in the MKG reasoning task.
\end{abstract}

\begin{IEEEkeywords}
Multi-modal knowledge graph, Multi-hop knowledge graph reasoning, Multi-modal fusion
\end{IEEEkeywords}

\section{Introduction}
Knowledge Graph (KG) is essentially a kind of graph structure with entities as nodes and relations as edges, and has received extensive attention in both data mining \cite{DBLP:conf/icde/DiYZC21} and knowledge engineering \cite{DBLP:conf/icde/ZhangYDC20} areas.
At present, large-scale KGs have achieved great success in assisting many applications, such as information retrieval \cite{DBLP:conf/acl/SunLXL18}, question answering \cite{DBLP:conf/sigir/KaiserRW21}, recommendation systems \cite{DBLP:conf/sigir/XianFMMZ19}\cite{DBLP:conf/ijcai/Zhao0Y0Z0X20} \cite{DBLP:journals/corr/abs-2004-02340}, etc. Nevertheless, most of traditional KGs only contain  structural data in the form of relation  triplets, i.e., (\emph{source entity}, \emph{relation}, \emph{target entity}), ignoring massive amounts of multi-modal data such as text and images in reality. To integrate more diverse knowledge in KGs, multi-modal Knowledge graph (MKG) has been proposed \cite{DBLP:conf/esws/LiuLGNOR19}  \cite{DBLP:journals/tamd/XieHLWYWS20}.  As the example presented in Fig. 1, a MKG not only contains the structural data, but also includes additional multi-modal auxiliary data  (i.e., texts and images), and it is more in line with the characteristics of real-world data compared with traditional KGs \cite{DBLP:conf/www/LuggenADC21}  \cite{DBLP:conf/cikm/SunCZWZZWZ20}  \cite{DBLP:journals/cee/XiongLLL21}.
Despite the abundant information contained by a MKG, it still suffers from the natural incompleteness of KGs,
 e.g., a triplet (\emph{Titanic}, \emph{Starred\_by}, \emph{Kate Winslet}) is missed in Fig. 1, which has significantly hindered the applications of MKGs  \cite{DBLP:conf/emnlp/LiC19}.

\begin{figure*}
 \centering 

  \label{Figure 2}
  \includegraphics[width=0.77\linewidth,height=8cm]{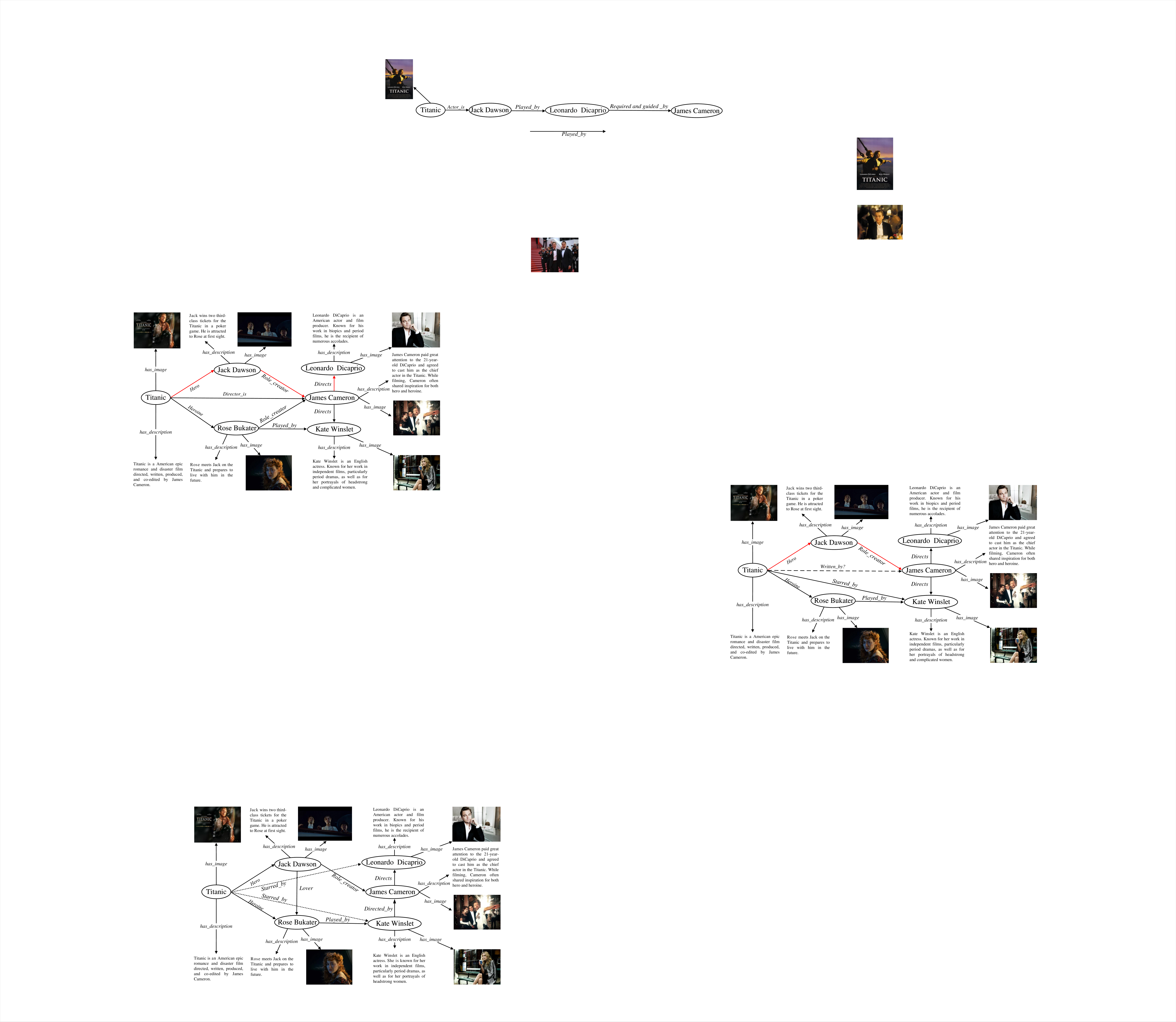}
 \hfill 
 \caption{
A small fragment of a multi-modal knowledge graph. The ellipse represents entities, the arrow refers to the relations and dotted arrow represents missing relation. When inferring (\emph{Titanic}, \emph{Starred\_{by}},  ?), the multi-modal auxiliary information such as the image or text  contained in each entity can provide reasoning clues for the reasoning model. To easily follow our work, we exemplify this multi-modal knowledge graph throughout the paper.}
\vspace{-15pt}
\end{figure*}

To solve the problem of natural incompleteness of KGs, various KG reasoning approaches have been proposed \cite{DBLP:conf/icde/LiGC20} \cite{DBLP:conf/icde/ZhangYSC19}\cite{DBLP:conf/iclr/DasDZVDKSM18} \cite{DBLP:conf/sigir/JiangGL21}. The key idea of these methods is to infer new knowledge by effectively integrating existing information in the graph \cite{DBLP:journals/eswa/ChenJX20}, and they mainly focus on traditional KGs without considering multi-modal knowledge. 
Different from these methods, some reasoning models \cite{DBLP:conf/aaai/XieLJLS16}  \cite{DBLP:conf/ijcai/XieLLS17}  \cite{DBLP:conf/ijcnn/WangLLZ19}  \cite{DBLP:conf/starsem/SergiehBGR18}
\cite{DBLP:conf/bigmm/ZuoFQZX18} \cite{DBLP:journals/ipm/TangCCW19} have been proposed to integrate the multi-modal knowledge on MKGs, but they are based on the model TransE \cite{DBLP:conf/nips/BordesUGWY13} that focuses on completing the single-hop reasoning only. Notably, the single-hop  reasoning models lack 
explainability and suffer from low reasoning performance since KGs have the most inferred potential knowledge within multiple hops \cite{DBLP:conf/iclr/DasDZVDKSM18}. Accordingly, there is another stream of KG reasoning methods focusing on multi-hop reasoning. 
A representative method in this stream is reinforcement learning (RL)-based multi-hop reasoning owing to its ability to leverage the symbolic combination and transmission of relations in KGs \cite{DBLP:conf/ijcai/WanP00H20}, which makes the whole reasoning process explainable \cite{DBLP:conf/dasfaa/WangYWHGN19}. For example, by connecting (\emph{Titanic}, \emph{Heroine}, \emph{Rose Bukater}) and  (\emph{Rose Bukater}, \emph{Played\_{by}}, \emph{Kate Winslet}), RL-based reasoning models obtain a missing triplet (\emph{Titanic}, \emph{Starred\_by}, \emph{Kate Winslet}). It has been proven that  RL-based multi-hop KG reasoning models have not only semantic explainability, but also higher reasoning performance than single-hop reasoning models \cite{DBLP:conf/emnlp/LinSX18}  \cite{DBLP:journals/tnn/JiPCMY22}\cite{DBLP:conf/cikm/Zheng0Z0F021}, which motivates our study to focus on multi-hop reasoning in MKGs. Note that, the existing multi-hop reasoning methods in KG area have not integrated the multi-modal information so far. An intuitive solution to conduct multi-hop reasoning in MKGs is to extend the existing multi-hop reasoning methods in traditional KGs to include the multi-modal information. However, the following two  \emph{challenges} make the direct extension ineffective.

The \emph{challenge \uppercase\expandafter{\romannumeral1}} is the lack of a fine-grained multi-modal information exploiting method in KG reasoning area. Several multi-modal studies have demonstrated that fine-grained features are beneficial for obtaining accurate results in reasoning tasks \cite{DBLP:conf/cvpr/Yu0CT019, DBLP:conf/aaai/LiSGLH0G19}. Typically, most of the existing MKG reasoning methods learn separate attention distributions for only one modality information (e.g., texts or images) apart from the structure information. Despite the accomplishment that the above methods have achieved, they still remain some unsolved problems. Firstly, these methods cannot simultaneously learn the visual attention and textual attention from fine-grained representations of inter-modal interactions \cite{DBLP:journals/ipm/TangCCW19}. By way of illustration, in Fig. 1, the potential triplet (\emph{Titanic}, \emph{Starred\_{by}}, \emph{Leonardo Dicaprio}) cannot be inferred until the images and text descriptions in the 3-hop path ``Titanic$\stackrel{Hero}{\longrightarrow}$ Jack Dawson $\stackrel{Role\_{creator}}{\longrightarrow}$ James Cameron $\stackrel{Directs}{\longrightarrow}$ Leonardo Dicaprio'' are simultaneously considered.  Secondly, existing multi-modal reasoning methods ignore intra-modal interactions that have been shown to be the key of fine-grained learning for many unimodal learning tasks \cite{DBLP:conf/nips/VaswaniSPUJGKP17, DBLP:conf/cvpr/0004GGH18}. For example,  we can infer that ``Titanic" is a love movie about two people embracing each other only from the image of the entity ``Titanic". Last but not least, some irrelevant noise (e.g., black background in images) and redundant noise (compared with the image of Rose Bukater, the image of Kate Winslet is highly similar and contains less useful information) impair the robustness and generalization of the model \cite{DBLP:journals/cee/ChandrashekarS14, DBLP:journals/tcyb/LiWR22}. How to learn fine-grained knowledge by simultaneously solving the above problems is non-trivial in MKG reasoning area.

The \emph{challenge \uppercase\expandafter{\romannumeral2}} lies in directly extending the RL-based KG reasoning method to MKG reasoning easily generates some wrong reasoning paths and degrades the reasoning performance. This is because the introduction of multi-modal auxiliary data further exacerbates the sparse reward problem, which leads to the decision bias of reinforcement learning \cite{DBLP:conf/sigdial/ZhangZY18,  DBLP:conf/emnlp/MisraLA17,  DBLP:conf/nips/HuYGTL19,  DBLP:conf/ijcai/ChaplotLSPB20}.  In fact, most RL-based KG reasoning methods have suffered from the problem of sparse reward on traditional KGs, which is manifested in the lack of feedback rewards and blind reasoning in most states \cite{DBLP:conf/emnlp/LinSX18}. Some KG reasoning methods have tried to alleviate this problem, but they still have the following technical limitations: (1) The general designed principles of density, exploration and restraint are not fully considered in the reward function, which leads to failure to converge in  the late phase of training \cite{DBLP:conf/nips/DevidzeRKS21, DBLP:conf/aaaiss/Dewey14}. (2) Lack of reward balancing mechanism to prevent reasoning models from repeatedly grabbing local rewards  while ignoring the ultimate goal \cite{ren2021orientation, therrien2016effective}.  (3) Lack of a paradigm for perceiving and exploiting multi-modal features in reinforcement learning.

To deal with the above challenges, we propose a novel model entitled MMKGR (\textbf{M}ulti-hop \textbf{M}ulti-modal \textbf{K}nowledge \textbf{G}raph \textbf{R}easoning). The main difference between our model and existing ones is that MMKGR not only effectively extracts and utilizes multi-modal auxiliary features, but also completes multi-hop reasoning in MKGs. Specifically,  the model contains the following two components. (1) To solve the \emph{challenge \uppercase\expandafter{\romannumeral1}}, a unified gate-attention network is designed to generate multi-modal complementary features with sufficient interactions and less noise. Its attention-fusion module extracts fine-grained  multi-modal features to complete inter-modal attention interactions (i.e., co-attention
across different modalities) and intra-modal attention interactions (i.e., self-attention within each modality) simultaneously. An irrelevance-filtration module of this network further filters out irrelevant features and outputs more reliable multi-modal complementary features. Note that, the unified gate-attention network  simultaneously aggregates low-noise information to obtain the triple query-related fine-grained representation from intra-modality and inter-modality.
(2) To solve the \emph{challenge \uppercase\expandafter{\romannumeral2}}, a complementary feature-aware RL method is proposed to predict the missing elements by performing the multi-hop reasoning process. More precisely, a carefully designed 3D reward mechanism, which includes \textit{Destination reward}, \textit{Distance reward} and \textit{Diverse reward}, is proposed in MMKGR. To sum up, this study makes the following contributions.
\begin{itemize}
	\item To the best of our knowledge, we are the first to investigate the problem of \textit{how to effectively leverage multi-modal auxiliary features to conduct multi-hop reasoning in KG area}, and this study provides a new perspective on KG reasoning.
	\item To resolve the above problem, we propose a novel model MMKGR,  which contains a unified gate-attention network that builds sufficient attention interactions with less noise, and a complementary feature-aware RL method that is designed to alleviate the problem of sparse rewards and conduct multi-hop reasoning in MKGs.
	\item Extensive experiments on two benchmark datasets have been conducted, and the results demonstrate better  performance of MMKGR against  state-of-the-art baselines.
\end{itemize}
	
The rest of the paper is organized as follows. Firstly, we review the related work in Section \uppercase \expandafter{\romannumeral2}. Then, we introduce the preliminary and formulate the problem in Section \uppercase \expandafter{\romannumeral3}. Section \uppercase \expandafter{\romannumeral4} presents MMKGR to complete the multi-hop reasoning in the MKG. Experiments are conducted in Section \uppercase \expandafter{\romannumeral5}, which is followed by the conclusion and future work in Section \uppercase \expandafter{\romannumeral6}.

\section{Related Work}
\subsection{Multi-modal Knowledge Graph}
A KG is essentially a structured semantic network composed of entities and relations  \cite{DBLP:conf/nips/ShenCHGG18} \cite{DBLP:conf/ijcai/TamWTYH17}. At present, the actual internet data show multi-modal characteristics \cite{DBLP:journals/tnn/JiPCMY22}. MKGs are proposed to integrate multi-modal data in KGs \cite{DBLP:conf/emnlp/PezeshkpourC018}  \cite{DBLP:conf/starsem/SergiehBGR18}. A  MKG is composed of structural data (i.e., relation triplets), and multi-modal auxiliary data (i.e., texts and images)  \cite{DBLP:conf/esws/LiuLGNOR19}.

The multi-modal auxiliary data associated with early MKGs present the singularity. For example, as a precedent for MKGs, the entities of IMGpedia \cite{DBLP:conf/semweb/FerradaBH17} come from a specific KG (i.e., DBpedia), and the multi-modal auxiliary data only contain images. Similar studies \cite{DBLP:conf/esws/LiuLGNOR19} \cite{DBLP:conf/emnlp/PezeshkpourC018} expand the existing KGs WN-9 and FB15K respectively, only adding images for each entity to further explain them. Although the above MKGs connect structural entities with images, they do not consider the diversity of images. To solve this problem, Richpedia \cite{DBLP:journals/bdr/WangWQZ20} filters out similar images to ensure diversity. There are also MKGs that only contain textual descriptions. One representation is FB20K \cite{DBLP:conf/aaai/XieLJLS16} that only textual descriptions are added to each entity. Although FB-Des \cite{DBLP:journals/ipm/TangCCW19} adds textual descriptions and  hierarchical types for each entity on the basis of FB15k-237, the auxiliary data of this MKG is still singular. 
To expand the auxiliary data with one modality, two datasets WN9-IMG-TXT and FB-IMG-TXT simultaneously add a number of textual descriptions and images to each entity, aiming to further enhance the data diversity of the MKGs \cite{DBLP:conf/starsem/SergiehBGR18}. 
While these MKGs add a large amount of multi-modal auxiliary data, they also generate redundant and irrelevant data, which limits the performance of multi-modal fusion.

\subsection{Fusion Strategies for Multi-modal Learning}
Early multi-modal studies only fuse the global features of all modalities through vector concatenation. The limitation of this method is that the multi-modal noise affects the extraction of key features \cite{DBLP:conf/emnlp/FukuiPYRDR16}. Thus, some multi-modal studies  \cite{DBLP:conf/cvpr/YangHGDS16}   \cite{DBLP:conf/cvpr/LuXPS17}  \cite{DBLP:conf/ijcai/XieLLS17} adopt conventional attention model to extract important features of auxiliary modalities. Compared with earlier fusion methods, the conventional attention mechanism aggregates essential information to obtain the key local representation \cite{DBLP:conf/icml/XuBKCCSZB15}.
Furthermore, considering that conventional attention mechanisms cannot perform feature interactions in all modalities at the same time,
some studies propose co-attention models to simultaneously assign and  aggregate essential information for all modalities \cite{DBLP:journals/tnn/YuYXFT18}  \cite{DBLP:conf/aaai/0001FLH18}. Although co-attention is extended to learn all modalities at the same time, these models, like conventional attention mechanism, still learn coarse interactions among all modalities.
To address the problem of insufficient multi-modal interactions, MCAN \cite{DBLP:conf/cvpr/Yu0CT019} and PSAC \cite{DBLP:conf/aaai/LiSGLH0G19} apply self-attention mechanism \cite{DBLP:conf/nips/VaswaniSPUJGKP17} and co-attention to complete the intra-modal and inter-modal attention interactions. 
The above methods have sufficient interaction, but ignore the following details: (1) redundant and irrelevant features can impair the generalization and robustness of the model \cite{DBLP:journals/tcyb/LiWR22}; (2) only self-attention or co-attention is considered in the same training stage,  which limits the utilization of samples and the effective extraction of fine-grained features \cite{DBLP:conf/nips/VaswaniSPUJGKP17}. 
The inadequacy of existing methods motivates our multi-modal fusion goal to simultaneously complete the intra-modal and inter-modal interactions in a unified and low-noise way.

\begin{table}
  \caption{Summarization of Existing KG Reasoning Models.} 
  \label{table5} 
  \resizebox{0.95\hsize}{!}{ 
    \begin{tabular}{c|c|c}   
    \hline
    \diagbox{Type}{Models}{KGs}    & On traditional KGs &On multi-modal KGs\\  
    \hline

        &    TransE, TransD &IKRL, DKRL  \\
            Single-hop reasoning &    ComplEx, HolE  &TransAE, MTRL  \\
            &    DistMult, RESCAL  &KR-AMD, MKRL  \\
      \hline
        &     MINERVA, DeepPath  &  \\
            Multi-hop reasoning &    GaussianPath,  RLH  &\textbf{MMKGR}  \\
            &    GAATs,   NeuralLP  &   \\
    \hline
    \end{tabular}
  }
\vspace{-10pt}
\end{table}

\begin{figure*}
	\centering 
	
	\label{Figure 2}
	\includegraphics[width=0.94\linewidth,height=4.9cm]{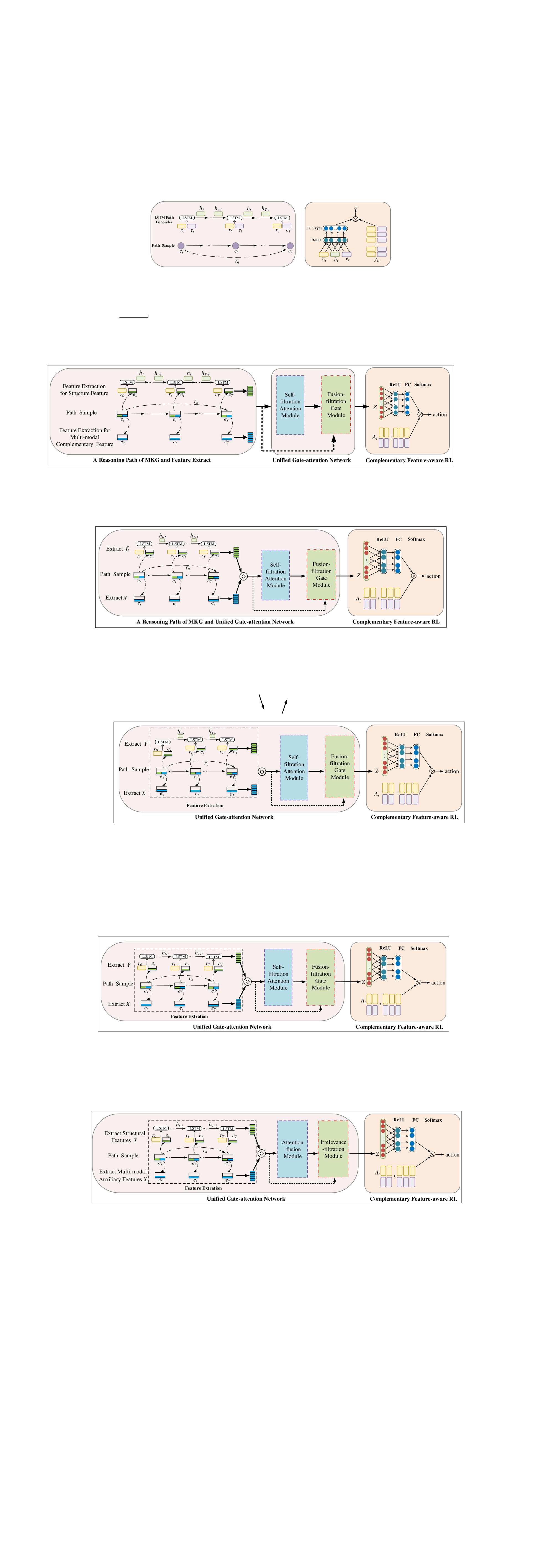}
	\hfill 
	\caption{
		Overview of MMKGR. The structural features $Y$ in green and the multi-modal auxiliary features $X$ in blue are obtained by feature extraction.
		These features at the reasoning step \emph{t} are fed to the unified gate-attention network to generate multi-modal auxiliary features $Z$ which are fed into the complementary feature-aware RL to predict action for the next reasoning step \emph{t+1}, until reaching the target entity $e_T$.
		}
		\vspace{-15pt}
\end{figure*}

\subsection{Knowledge Graph Reasoning}
Since KGs are inherently incomplete, KG reasoning technology that can synthesize the original knowledge and infer the missing knowledge is particularly important \cite{DBLP:journals/tnn/JiPCMY22} \cite{DBLP:journals/vldb/HungDTWAYZ17}. We summarize the existing knowledge graph reasoning models in Table \uppercase \expandafter{\romannumeral1}. A group of KG reasoning studies on traditional KGs aim at 
inferring missing elements by embedding-based methods \cite{DBLP:conf/AAAI/Stoica20}  \cite{DBLP:conf/nips/BordesUGWY13}  and deep learning \cite{DBLP:conf/aaai/DettmersMS018} \cite{DBLP:conf/esws/SchlichtkrullKB18}. For example, the embedding-based TransE \cite{DBLP:conf/nips/BordesUGWY13} learns vector representations of entities and relations by minimizing the heuristic self-supervised loss functions and then the learned vectors are used to predict the probability of correct triplets. However, all the above methods are only suitable for single-hop reasoning by modeling one-step relations
containing less information  \cite{DBLP:conf/iclr/DasDZVDKSM18}. Moreover, these methods cannot leverage the symbolic combination and transmission of relations in KGs \cite{DBLP:conf/ijcai/WanP00H20}.

Recent developments of KG reasoning on traditional KGs focus on the field of multi-hop reasoning based on RL.
The reasoning process of RL-based reasoning method is intuitive by exploiting the symbolic compositionality of multi-step relations containing more information  \cite{DBLP:conf/iclr/DasDZVDKSM18}. 
Typical RL-based methods include MINERVA  \cite{DBLP:conf/iclr/DasDZVDKSM18}, DeepPath \cite{DBLP:conf/aaai/WanD21}, RLH \cite{DBLP:conf/ijcai/WanP00H20}, GaussianPath \cite{DBLP:conf/aaai/WanD21},  etc. In addition, there are other multi-hop methods, e.g., rules-based NeuralLP \cite{DBLP:conf/nips/YangYC17}  and graph attention networks-based GAATs \cite{DBLP:journals/access/RocktaschelR17}. In spite of the superior performance of 
these existing multi-hop KG reasoning methods, they ignore the multi-modal data types, and cannot use multi-modal auxiliary data to complete reasoning. As presented in Table \uppercase \expandafter{\romannumeral1}, our MMKGR fills the gap for multi-hop reasoning on MKGs.

Focusing on single-hop reasoning on MKGs, some studies employ the conventional attention model or concatenation to fuse multi-modal features and then adopt TransE to infer missing elements, such as IKRL \cite{DBLP:conf/ijcai/XieLLS17}  \cite{DBLP:journals/tamd/XieHLWYWS20} and TransAE \cite{DBLP:conf/ijcnn/WangLLZ19}. Note that, Wang et al. have proved that the performance of TransAE on MKGs is better than that of the most traditional KG reasoning methods, such as TransE, RESCAL \cite{DBLP:conf/icml/NickelTK11}, ComplEx \cite{DBLP:conf/icml/TrouillonWRGB16}, HolE \cite{DBLP:conf/aaai/NickelRP16}, and DistMult \cite{DBLP:journals/corr/YangYHGD14a}.
KR-AMD \cite{DBLP:conf/bigmm/ZuoFQZX18} and MKRL \cite{DBLP:journals/ipm/TangCCW19} leverage textual data as part of auxiliary data to improve reasoning performance.
In addition, MTRL \cite{ DBLP:conf/starsem/SergiehBGR18} with the state-of-the-art performance performs single-hop reasoning by concatenating the features of relation triplets and multi-modal auxiliary features that comprehensively contain textual and visual features. 
However, the above studies cannot simultaneously learn visual and textual attention to fully understand the semantics of the two modalities. This coarse interaction causes ineffective  fusion of multi-modal features.

\section{Preliminary and Definition}

A	knowledge graph 
	\begin{math}
		\mathcal{G} = \{\mathcal{E}, \mathcal{R}, \mathcal{U}\} 
	\end{math} is a directed heterogeneous graph, where $\mathcal{E}$ is the set of entities and $\mathcal{R}$ is the set of relations.
	\begin{math}
		\mathcal{U} = \{ (\emph{$e_s$}, \emph{r}, \emph{$e_d$}) \mid \emph{$e_s$}, \emph{$e_d$} \in \mathcal{E}, \emph{r} \in \mathcal{R} \} 
	\end{math} is a set of relation triplets in $\mathcal{G}$, where $e_s$, $e_d$, and $r$ are a source entity, a target entity, and the relation between them, respectively. Relation triplets in knowledge graphs are structured \cite{DBLP:conf/nips/BordesUGWY13}, and the corresponding structural features are different from the features learned from auxiliary data (i.e., texts and images).
To better understand our methodology in this study, some definitions are introduced as follows.
	
\begin{defn}\emph{Multi-modal knowledge graph}.
	A multi-modal knowledge graph \begin{math}
	    \tilde{\mathcal{G}} = \{\tilde{\mathcal{E}}, \mathcal{R}, \tilde{\mathcal{U}}\} 
	\end{math} is an extension of the knowledge graph $\mathcal{G}$ by adding multi-modal auxiliary data, where each entity in $\tilde{\mathcal{E}}$ is attached to both structural (i.e., the relation triplets in knowledge graphs) and multi-modal data.
\end{defn}
	
 	\begin{defn}
		\label{def:activity}
			\emph{Multi-hop reasoning}. Given a query among three cases  $($\emph{$e_s$}, \emph{$r$}, ?$)$, $($\emph{$e_s$}, ? , \emph{$e_d$}$)$, $($? , \emph{$r$}, \emph{$e_d$}$)$, where ``?" represents the missing element, the goal of multi-hop reasoning is to predict the missing element through a reasoning path shorter or equal $k$ hops, where $k$ is an integer not less than 1.

		\emph{Example}: Given a triple query (\emph{Titanic},  \emph{Starred\_by}, ?), a 3-hop reasoning path is ``Titanic$\stackrel{Hero}{\longrightarrow}$ Jack Dawson $\stackrel{Role\_{creator}}{\longrightarrow}$ James Cameron $\stackrel{Directs}{\longrightarrow}$ Leonardo Dicaprio''.
	\end{defn}

		\begin{defn}
		\label{def:activity}
			\emph{Multi-modal auxiliary feature}. 
Multi-modal auxiliary feature $\emph{\textbf{x}}$ of each entity $e$ in entity set $\tilde{\mathcal{E}}$ is expressed as a vector, learned from multi-modal auxiliary data (text or image) and $\emph{\textbf{x}}$ $\in$ $\textbf{\emph{f}}_{t}$   
 $\circ$ $\textbf{\emph{f}}_{i}$, where ``$\circ$" represents a multi-modal fusion method. The textual feature vector  $\textbf{\emph{f}}_{t}$, image feature vector  $\textbf{\emph{f}}_{i}$, and  $\emph{\textbf{x}}$ are learned with some representation methods.
	\end{defn}

\newtheorem{problem}{Problem}
\begin{problem}
	This study aims to predict the missing entity or relation for a given triple query on the multi-modal knowledge graph. The problem formulation is given as:
	\begin{itemize}
		\item Input is a triple query (?, \emph{$r_q$}, $e_{d}$), ($e_{s}$, \emph{$r_q$}, ?) or ($e_{s}$, ?, $e_{d}$), where $r_q$ is a query relation, $e_{s}$ and $e_{d}$ are the source entity and target entity containing  structural feature and multi-modal auxiliary feature, and ``?" is the missing element.
		\item Output is a predicted entity or relation obtained by multi-hop reasoning.
	\end{itemize}
\end{problem}

	\begin{figure}[h]
	\begin{minipage}[t]{0.45\textwidth}
		\centering
		\includegraphics[width=\textwidth,height=9cm]{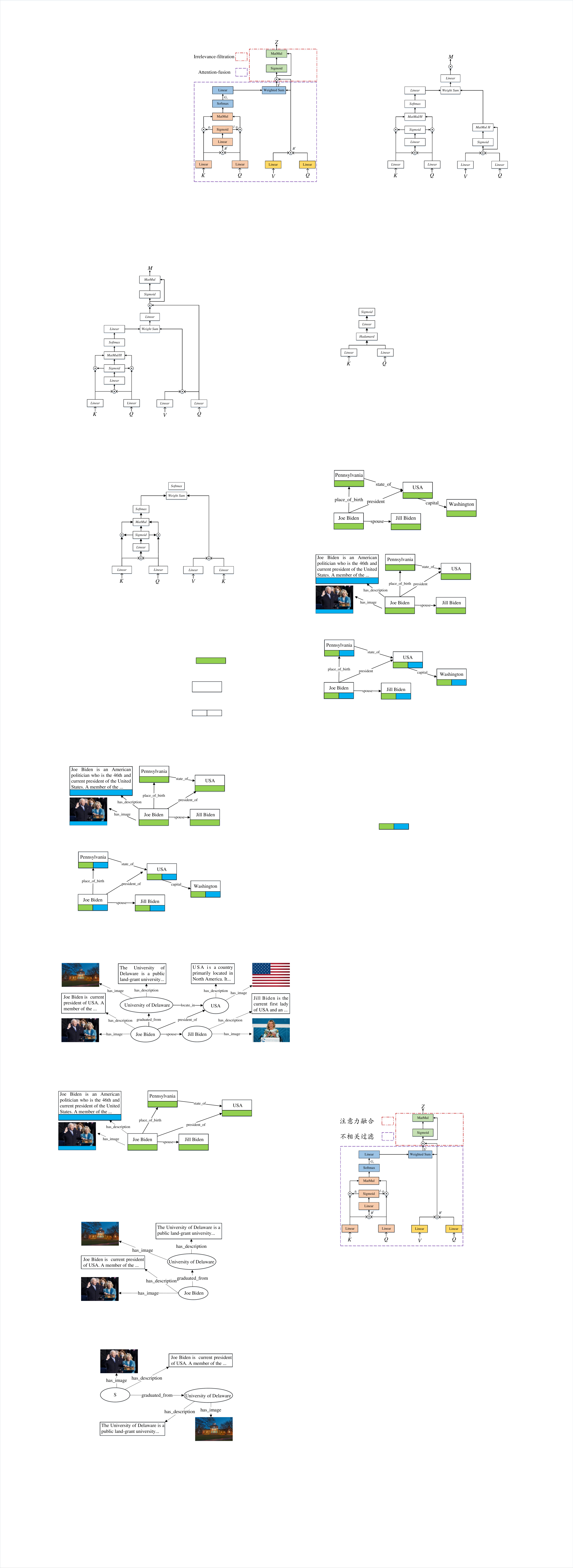}
		
	\end{minipage}
	
	\setlength{\belowcaptionskip}{-0.2cm}
	\caption{The detailed schematic diagram of the unified gate-attention network to generate multi-modal complementary features.}\label{FIGURE2}
	\vspace{-17pt}
\end{figure}
	
\section{Methodology}
\subsection{Overview of MMKGR}\label{4:Overview of THML}
Our proposed model MMKGR, the overview of which is shown in Fig. 2, contains two components: (1) a unified gate-attention network, which is designed to conduct sufficient attention interactions and filter noises to generate more effective and reliable multi-modal complementary features encoding relevant knowledge of all modalities; (2) a complementary feature-aware reinforcement learning framework, which is proposed to predict the missing elements in the multi-hop reasoning process with a carefully designed reward mechanism and the useful multi-modal complementary features.

\subsection{Unified Gate-attention Network}
The insufficient multi-modal interactions and noise interference in existing MKG reasoning methods severely limit the utilization of multi-modal data \cite{DBLP:conf/cvpr/Yu0CT019}  \cite{DBLP:conf/cvpr/NguyenO18}. To address the problem, we introduce a novel and unified gate-attention network to learn from multi-modal data in this subsection. Notably, this network is mainly inspired by (1) the significant impact of intra-modal attention interaction on fine-grained features \cite{DBLP:conf/nips/VaswaniSPUJGKP17}, and (2) gate network which effectively filters noise \cite{DBLP:conf/aaai/0001FLH18}. Based on this, the unified gate-attention network selects features of different modalities online, and simultaneously completes intra-modal and inter-modal attention interactions with noise robustness, which is technically different from both existing KG multi-modal data modelling methods \cite{DBLP:conf/aaai/XieLJLS16}  \cite{DBLP:conf/ijcai/XieLLS17}  \cite{DBLP:conf/ijcnn/WangLLZ19}  \cite{DBLP:conf/starsem/SergiehBGR18} \cite{DBLP:journals/ipm/TangCCW19} and general multi-modal data modelling approaches \cite{DBLP:conf/aaai/LiSGLH0G19} \cite{DBLP:conf/nips/VaswaniSPUJGKP17} that are not limited to KG area.

Specifically, the unified gate-attention network includes feature extraction, an attention-fusion module, and an irrelevance-filtration module. 
The extracted features of all modalities are fed into the attention-fusion module that fuses structural features and multi-modal auxiliary features together, by attending them with a carefully designed fine-grained attention scheme. Then, the irrelevance-filtration module discards irrelevant or even misleading information and generates noise robust multi-modal complementary features. Fig. 3 presents the schematic diagram of the unified gate-attention network, the details of which are illustrated as follows. 

\subsubsection{Feature Extraction}
(1) Structural features with \emph{$d_s$} dimensions are initialized from all entities and relations by using the TransE algorithm \cite{DBLP:conf/nips/BordesUGWY13}. The source entity $e_s$ and query relation $r_q$ are represented as the dense vector embedding $\textbf{e}_{s}$ and $\textbf{r}_{q}$ respectively. In addition, the history of  reasoning path that consists of the visited entities and relations is defined as $h_t$ = ($e_s$, $r_0$, $e_1$, $r_1$,...,$e_t$). Next, we leverage LSTM to integrate the vector of history information $\textbf{h}_t$  with \emph{$d_s$} dimensions into structural features.
Given the query in our multi-hop reasoning process, the structural features \textbf{y} are defined as,
\begin{equation}
	\textbf{y} = [\textbf{e}_{s};\textbf{h}_t;\textbf{r}_{q}]
\end{equation}
\begin{equation}
	Y = [\textbf{y}_1,\textbf{y}_2,...,\textbf{y}_{m}]
\end{equation}
where \begin{math}
	Y \in \mathcal{R}^{m \times {d_y}}
\end{math} represents a group of structural features, 
\emph{m} and \emph{$d_{y}$} are the number of entities and the dimension of the features in this triple query, respectively.
(2) To initialize image features $\textbf{f}_{i}$, we extract a \emph{$d_i$}-dimensional vector of the last fully-connected layer before the softmax in VGG model \cite{DBLP:conf/bmvc/ChatfieldSVZ14}.
(3) Textual features $\textbf{f}_{t}$ are initialized by the word2vec framework \cite{DBLP:journals/corr/abs-1301-3781} and expressed as a \emph{$d_t$}-dimensional vector.
To flexibly add multi-modal auxiliary features, 
we concatenate the above two groups of features on rows to form the multi-modal auxiliary features \textbf{x},
\begin{equation}
	\textbf{x} = [\textbf{f}_{t}W_{t};\textbf{f}_{i}W_{i}]
\end{equation}
\begin{equation}
	X = [\textbf{x}_{1},\textbf{x}_{2},...,\textbf{x}_{m}]
\end{equation}
where $\emph{$W_t$} \in \mathcal{R}^{d_t \times {d_x}/2}$, $\emph{$W_i$} \in \mathcal{R}^{d_i \times {d_x}/2}$, and \begin{math}
	X \in \mathcal{R}^{m \times {d_x}}
\end{math} represents a group of multi-modal auxiliary features,
\emph{$d_{x}$} is the dimension of the feature.

\subsubsection{Attention-fusion Module}

To obtain the complementary features for reinforcement learning, we need to fuse the structural features $Y$ and multi-modal auxiliary features $X$ generated in feature extraction. 
However, redundant features tend to have a negative impact on the prediction during the multi-modal fusion \cite{DBLP:conf/aaai/0001FLH18}. 	Specifically, redundant features are either shifted versions of the features related to the triple query or very similar with little or no variations  \cite{DBLP:journals/corr/abs-1802-07653}, which can amplify the negative effects of noise  \cite{DBLP:journals/tbe/LiLSLLZHZYZ019}. 
For example,  suppose we have  ``Titanic$\stackrel{Heroine}{\longrightarrow}$ Rose Bukater $\stackrel{Played\_{by}}{\longrightarrow}$ Kate Winslet'' and the queries are about the movie Titanic, the features of an image containing Rose Bukater provide related information for the queries while the features coming from another very similar image or the images containing Kate Winslet (Rose’s player) in other movies are regarded as redundant features. These redundant features add computational complexity and cause collinearity problems \cite{DBLP:conf/ijcnn/WangLLZ19} \cite{DBLP:journals/ipm/TangCCW19}. Consequently, we propose the attention-fusion module that is located in the lower area of Fig. 3, with the goal of fusing the structural features and multi-modal auxiliary features effectively.

Specifically, we first utilize linear functions to generate the queries \emph{Q}, keys \emph{K}, and values \emph{V} of the attention mechanism,
\begin{equation}
	Q = XW_{q}, K= YW_{k}, V= YW_{v}
\end{equation}
where 
\begin{math}
	W_{q}\in\mathcal{R}^{d_x \times d}, W_{k}, W_{v}\in\mathcal{R}^{d_y \times d},\end{math}  and \begin{math} Q, K, V \in \mathcal{R}^{m \times d}
\end{math} have the same shape. Then, the joint representation $B^l$ of \emph{Q} and \emph{K} is learned based on MLB pooling method \cite{DBLP:conf/iclr/KimOLKHZ17}, inspired by the recent successes of it in fine-grained multi-modal fusion,
\begin{equation}
      B^l = KW_{k}^{l} \odot QW_{q}^{l}
\end{equation} Similarly, we can generate the joint representation $B^r$ of \emph{V} and \emph{Q} with the following equation,
\begin{equation}
	B^r = VW_{v}^{r} \odot QW_{q}^{r}
\end{equation}
where $W_{k}^{l}, W_{q}^{l}, W_{v}^{r}, W_{q}^{r} \in \mathcal{R}^{d \times j}$ are embedding matrices, and $\odot$ is Hadamard product.

Next, the filtration gate $g_t$ applied to different feature vectors is defined as,
\begin{equation}
	 g_t = \sigma(B^lW_{m}) 
\end{equation}
where $W_{m} \in \mathcal{R}^{j \times d}$ is an embedding matrix and $\sigma$ denotes the sigmoid activation. Based on the filtration gate $g_t$, we can filter out the redundant features generated during fusion and obtain a new representation with following probability distributions,
\begin{equation}
	G_s = softmax((g_t \odot K) ((1-g_t)\odot Q))
\end{equation}
where $g_t$ and $1-g_t$ are used to trade off how many structural features and multi-modal auxiliary features are fused.

Finally, our attention-fusion module generates the attended features \emph{$\hat{V}$}=\{${\textbf{v}_i}\}_{i=1}^{m}$ by accumulating the enhanced bilinear values of structural features and multi-modal auxiliary features,
\begin{equation}
	\emph{$\hat{V}$} = \sum \nolimits_{i=1}^m (G_sW_{g}^{l})B_i^r
\end{equation}
where $W_{g}^{l}\in \mathcal{R}^{d \times 1}$, and $\textbf{v}_i \in \mathcal{R}^{1 \times j}$ denotes a row of the attended  features $\emph{$\hat{V}$} \in \mathcal{R}^{m \times j}$, feature vector $B_{i}^{r}\in \mathcal{R}^{1 \times j}$ is a row of the embedding matrix $B^{r}$.

By designing the attention-fusion module, we can complete the intra-modal and inter-modal feature interactions in a unified manner at the same time. This is because the inputs of this module are pairs from structural features and multi-modal auxiliary features, where each vector of a pair may be learned from the same modality or different ones.

\subsubsection{Irrelevance-filtration Module}
To further improve the robustness of the model, we design an irrelevance-filtration module, which is located in the upper area of Fig. 3. The attended features $\hat{V}$ obtained by attention-fusion module may contain irrelevant features \cite{DBLP:conf/iccv/HuangWCW19}. Specifically, irrelevant features are irrelevant to the triple query in the reasoning process. Since the attention mechanism assigns weights to all features, these features tend to participate in model computation and mislead the reasoning policy \cite{DBLP:conf/cvpr/RahmanCSC21}. For example, the features from the black background of images in Fig. 1 are  regarded as irrelevant features.
This motivates our model to weight more on the most related complementary features and dynamically filter irrelevant ones. This is achieved by a well designed irrelevance-filtration gate function. The output of this gate is a scalar, the value range of which is [0,1]. The  multi-modal complementary features $\emph{Z}$ are obtained as follows,
\begin{equation}
	G_f = \sigma(B^r\odot\hat{V})
\end{equation}
\begin{equation}
	Z = G_f(B^r\odot\hat{V})
\end{equation}
where $\sigma$ and $G_f$ denote
the sigmoid activation function and  irrelevance-filtration gate, respectively.

\subsection{Complementary Feature-aware Reinforcement Learning}\label{Complementary Feature-aware Reinforcement Learning}
The existing KG reasoning  methods based on RL are not suitable to be directly applied to reasoning in MKGs due to the dilemma of sparse rewards \cite{DBLP:conf/emnlp/LinSX18}. Sparse rewards (i.e., the agent cannot get enough rewards in a short period of time) are more likely to generate wrong reasoning paths, and the multi-modal auxiliary features of entities on these paths aggravate the introduction of noise, thereby further affecting the performance of reasoning. To solve this problem, we propose a novel reward mechanism in this subsection.   Compared with existing RL frameworks, the main technical difference of this work lies in the following two points. (1) A carefully-designed 3D reward mechanism that combines reward design principles with domain-knowledge of KGR is proposed to eliminate reward sparsity. (2) A novel method,  where the policy function is used as a multi-modal perception interface, is first introduced in RL to fully utilize the multi-modal features.

Our proposed model MMKGR  transforms the process of reasoning about missing elements into the Markov Decision Process (MDP) where the goal is to take a sequence of the optimal decisions (choices of relation paths) to maximize the expected reward (reaching the correct entity). The process of RL-based method is as intuitive as “go for a walk”, which naturally forms an explainable provenance for multi-hop reasoning. Thus, MMKGR trains an agent to interact with the sample environment of MKGs in the form of a \emph{4}-tuple of MDP (States, Actions, Transition, Rewards). 

States:\quad State of the RL-based agent corresponds to a set that includes some elements in MKGs. Formally, each state at reasoning step \emph{t} is denoted as \emph{$s_t$}= (\emph{$e_t$}, (\emph{$e_s$}, \emph{$r_q$}), $\mathcal{N}_t$, $\mathcal{E}_t$) $\in$ $\mathcal{S}$, where $\mathcal{S}$ is a state space and \emph{$e_t$} is the entity that is accessed at step \emph{t}. \emph{$e_s$} and \emph{$r_q$} are the source entity and query relation respectively. 
In addition, to ensure the design of a complete state, we also consider the set of neighboring entities $\mathcal{N}_t$ and all edges $\mathcal{E}_t$ connected to \emph{$e_t$}.

Actions:\quad Action space $\mathcal{A}_t$ for the given \emph{$s_t$} includes the set of valid actions \emph{$A_t$} at step \emph{t} and \emph{STOP} action.  \emph{$A_t$} is the set of outgoing edges connected to \emph{$e_t$}, and extends the path until  reaching next entity in MKGs. Formally, \emph{$A_t$} is 
expressed as \emph{$A_t$} = \{(\emph{$r_{t+1}$}, \emph{$e_{t+1}$})$\mid$ (\emph{$e_t$}, \emph{$r_{t+1}$}, \emph{$e_{t+1}$}) $\in$  $\mathcal{G}$\}. To avoid infinite unroll in the reasoning process, the \emph{STOP} action is executed when the reasoning step \emph{t} increases to the maximum step \emph{T}.

Transition:\quad  Transition function  $\mathcal{P}_r$ is set to map current state \emph{$s_t$} to a next state \emph{$s_{t+1}$}. Formally, $\mathcal{P}_r$: $\mathcal{S}$ $\times$ $\mathcal{A}$ $\rightarrow$ $\mathcal{S}$ is defined as $\mathcal{P}_r$ (\emph{$s_t$}, \emph{$A_t$}) = $\mathcal{P}_r$ (\emph{$e_t$}, (\emph{$e_s$}, \emph{$r_q$}), $\mathcal{N}_t$, $\mathcal{E}_t$, \emph{$A_t$}). When the search is unrolled to a fixed step \emph{T}, \emph{$s_t$} is denoted as \emph{$s_T$}.

Rewards:\quad Reward is a key factor affecting the performance of all RL-based models \cite{DBLP:journals/ai/SilverSPS21}.  Since directly extending the RL-based
KG reasoning method to MKG reasoning will exacerbate sparse rewards and degrades reasoning performance, we propose a \textbf{3D} reward mechanism (\textbf{Destination reward}, \textbf{Distance reward}, and \textbf{Diverse reward}) inspired by  \cite{DBLP:conf/emnlp/XiongHW17}.
According to the general principles of reward design, we employ reward shaping to maintain density of reward and introduce domain-knowledge (i.e. some inter-path and intra-path information in KGR) to present constraints and explorations of reward. Note that, the 3D reward mechanism not only preserves the three general principles of reward design, but also extracts intra-path and inter-path influences from the KGR domain, which ensures the completeness of reward design.
(1) \textbf{Destination reward}. 
When the agent unrolls to the maximum step \emph{T} or reaches \emph{$e_T$}, the agent receives the returned destination reward. For existing RL-based methods, the destination reward is 1 if \emph{$e_T$} is the ground truth target entity \emph{$e_d$}.  Otherwise, the destination reward is 0. We argue that this setting makes the rewards more sparse with the increase of the length of the reasoning path. For example,  in Fig. 1, if  (\emph{Titanic}, \emph{Starred\_by}, \emph{Kate Winslet}) or  (\emph{Titanic}, \emph{Starred\_by}, \emph{Leonardo Dicaprio})  still are not inferred at the maximum step \emph{T}, the above methods fail to obtain an effective reward (i.e., the value of destination reward is 0) and learn slowly or even ineffectively from multi-modal data. To alleviate the problem, we use the reward shaping trick \cite{DBLP:conf/emnlp/LinSX18} to design the reward when \emph{$e_d$} is not reached,
\begin{equation}
 \emph{R}_{destination}=
\begin{cases}
\qquad 1& \text{\emph{$e_T$} = \emph{$e_d$}}\\
\emph{l} (\emph{$e_s$}, \emph{$r_q$}, \emph{$e_T$})& \text{\emph{$e_T$} $\neq$ \emph{$e_d$}}
\end{cases}
\end{equation}
where \emph{l} is a score function using 
ConvE \cite{DBLP:conf/aaai/DettmersMS018} and is used to evaluate the probability over (\emph{$e_s$}, \emph{$r_q$}, \emph{$e_T$}).  
(2) \textbf{Distance reward}. 
The degree of sparse rewards tends to be positively correlated with the length of the reasoning path. As shown in Fig. 1, the triplet (\emph{Titanic}, \emph{Starred\_by}, \emph{Kate Winslet}) obtained through 2-hop reasoning path gets the terminal reward faster than (\emph{Titanic}, \emph{Starred\_by}, \emph{Leonardo Dicaprio}) obtained through 3-hop reasoning path. In fact, when the number of hops exceeds 3, the reasoning performance may be at risk of degradation  \cite{DBLP:conf/iclr/DasDZVDKSM18}. Therefore, we use the  $\emph{R}_{\emph{distance}}$ as part of the reward function to alleviate sparse rewards within a shorter path,
\begin{equation}
	\emph{R}_{\emph{distance}}=
	\begin{cases}
		\qquad \frac{1}{\emph{k}}& \text{\emph{$k$}} \leqslant \text{\emph{$3$}}\\ \qquad
		-\frac{1}{\emph{k}^2}& \text{\emph{$k$}} > \text{\emph{$3$}}
	\end{cases}
\end{equation}
where \emph{k} is the number of hops for the agent at step $t$, 3 is a threshold of \emph{k}. 
(3) \textbf{Diverse reward}. The lack of exploration further exacerbates sparse rewards. For example, in Fig. 1, given a reasoning task (\emph{Titanic}, \emph{Starred\_by}, ?), the agent can successfully complete reasoning via ``Titanic$\stackrel{Heroine}{\longrightarrow}$ Rose Bukater $\stackrel{Played\_{by}}{\longrightarrow}$ Kate Winslet''. The path found early will be biased, which limits exploration of novel paths. Thus, $\emph{R}_{\emph{diversity}}$ based on the Gaussian kernel is encouraged to
explore a diverse set of paths and 
prevents the agent from falling into the locally optimal path that negatively impacts on rewards,
\begin{equation}
	\emph{R}_{\emph{diversity}}= -\left| \frac{1}{V} \right| exp(-\frac{\lVert \textbf{p}-\textbf{p}_i \rVert}{2u^2})
\end{equation}
where \textbf{p} is the embedding of the relation path, \emph{V} is the known number of reasoning paths, and \emph{u} is a hyper parameter. In all, to balance the impact of each component of the 3D reward on reasoning performance, the reward \emph{R} is the linear combination of the following reward functions, which is defined as,
\begin{equation}
      \emph{R} = \lambda_{1}\emph{R}_{destination}+\lambda_{2}\emph{R}_{\emph{distance}}+\lambda_{3}\emph{R}_{\emph{diversity}}
\end{equation}
where $\lambda_i$ is a discount factor and $\lambda_1+\lambda_2+\lambda_3=1$.

\begin{table}
	\centering
	\caption{Statistics of the experimental datasets.}
	\begin{tabular}{llllll}
		\hline
		Dataset     & \#Ent  & \#Rel        & \#Train &\#Valid &\#Test  \\
		\hline
		WN9-IMG-TXT       & 6,555  & 9     & 11,747& 1,337 & 1,319\\
		FB-IMG-TXT       & 11,757  & 1,231      & 285,850 & 29,580 & 34,863\\
		
		\hline
	\end{tabular}
	
	\label{tab:plain}
	\vspace{-0.5cm}
\end{table}

Policy Network:\quad We design a policy network to drive the interaction between the explicitly defined MDP and MKGs. This policy network inputs the multi-modal complementary features and outputs the next action with the highest executable probability. In fact, the policy network is a feed-forward neural network. 
Specifically, the feed-forward neural network $\pi$ selects the next action in the action space $\mathcal{A}_t$ with the maximum probability, and $\pi$ is defined as follow,
\begin{equation}
    \emph{$\pi$}_\theta(\emph{a}_t|\emph{s}_t) = softmax (\textbf{A}_{t}(\textbf{W}_{2}\text{ReLu}(Z)))
\end{equation}
where $\mathcal{A}_t$ is encoded as $\textbf{A}_{t}$ by stacking the embedding of all actions. Multi-modal complementary features $Z$ are obtained from the unified gate-attention network. 

To maximize the accumulated rewards of our model and obtain the optimal policy, the objective function is as follow,
\begin{equation}
\emph{J}(\theta) = E_{(e_s,r,e_d) \sim \mathcal{G}_f}E_{a_1,...,a_{T}\sim {\emph{$\pi$}_\theta}}[\emph{R} (S_T\mid\emph{e}_s,r)]
\end{equation}

Finally, the stochastic gradient is used to conduct optimization.
\vspace{-5pt}
\begin{equation}
	\nabla_{\theta} \emph{J}(\theta) =\nabla_{\theta} \sum_{t=1}^TR(S_T|e_s,r)log\pi_{\theta}(a_t|s_t)
\end{equation}

\begin{table*}
	\centering
	\caption{Results of Entity Link Prediction on Two Multi-modal Knowledge Graphs.}
	\setlength{\tabcolsep}{3.3mm}{
		\begin{tabular}{c|cccc|cccc}
			\toprule
			&  \multicolumn{4}{c|}{WN9-IMG-TXT} & \multicolumn{4}{c}{FB-IMG-TXT} \\ 
			Model         & MRR    & Hits@1 & Hits@5    &Hits@10    & MRR        & Hits@1   & Hits@5    & Hits@10    \\ \midrule
			MTRL     & 48.3     & 45.6  & 69.8 & 83.8 & 25.2  & 21.3 & 32.4 & 47.2 \\
			NeuralLP     & 41.3     & 36.5 & 60.4  & 80.7 & 22.1  & 18.0 & 25.7  & 34.8 \\
			MINERVA     & 47.2     & 43.1 & 65.6  & 83.2 & 23.4  & 19.2 & 30.6  & 43.9 \\
			FIRE     & 56.4     & 52.8 & 77.6  & 86.8& 42.8  & 37.9 & 49.5 & 57.1 \\
			GAATs     & 58.2     & 54.6 & 79.4  & 87.7 & 45.4  & 41.2 & 54.3  & 61.8 \\
			RLH     &{\underline {62.4}}     &{\underline{58.3}}    & {\underline{81.3}} &{\underline{89.4}}   &{\underline{50.6}}   &{\underline{44.5}}  & {\underline{60.2}}  &{\underline{68.4}} \\
			MMKGR    & {\bf80.2}     & {\bf73.6} & {\bf 87.8} & {\bf 92.8} & {\bf71.3}  & {\bf65.8}  & {\bf77.5} & {\bf82.6} \\
			\bottomrule
			Improv.    &17.8\%     & 15.3\%  & 6.5\% &  3.4\% & 20.7\%  & 21.3\% & 17.3\% & 14.2\% \\
			\bottomrule
		\end{tabular}
	}
	\label{tab:whoiswho}
	\vspace{-5pt}
\end{table*}

\begin{table*}
\centering
  \caption{MAP of Relation Link Prediction on WN9-IMG-TXT and FB-IMG-TXT.} 
  \label{table5} 
  \setlength{\tabcolsep}{2.2mm}{
    \begin{tabular}{c|cccccc|c}   
    \hline
    \textbf{Tasks}  &MTRL &NeuralLP &MINERVA &FIRE &GAATs &RLH &MMKGR\\ 
    \hline

      has\_part  &65.6 &55.2 &64.7 &     75.9& 77.8 &{\underline{84.6}}&\textbf{98.2}\\
  
      derivationally\_related  &62.7 &53.3 &60.5 & 73.6 & 74.4&{\underline{80.7}}&\textbf{95.4}\\
    
    domain\_topic  &60.2 &48.7 &58.3 & 70.1 & 70.5&{\underline{78.6}}&\textbf{92.5}\\
   
    ....  & & &  & & & \\
      \hline
    Overall  &63.8 &54.3 &61.6 & 74.0&75.2 &{\underline{83.4}}&\textbf{97.1}\\
    \hline
  
      place\_founded  &41.5 &36.7 &40.2 &     63.5& 67.3 &{\underline{70.4}}&\textbf{88.3}\\
  
      producer\_type  &50.3 &43.6 &46.0 & 65.9 & 73.1&{\underline{79.4}}&\textbf{93.2}\\
    
    registering\_agency  &45.7 &42.4 &42.9 & 64.7 & 69.8&{\underline{72.7}}&\textbf{91.6}\\
   
    ....  & & &  & & & \\
      \hline
    Overall  &48.7 &43.1 &45.4 & 67.8&70.4 &{\underline{74.6}}&\textbf{93.6}\\
    \hline
    \end{tabular}
  }
  \vspace{-0.45cm}
\end{table*}

\section{Experiments}

\subsection{Experimental Setup}
\subsubsection{Datasets} 
We study the performance of  MMKGR on two public  datasets$\footnote{https://public.ukp.informatik.tu-darmstadt.de/starsem18-multimodalKB}$, i.e., WN9-IMG-TXT and FB-IMG-TXT. Both of them are MKGs widely adopted by existing reasoning studies \cite{DBLP:journals/tamd/XieHLWYWS20} \cite{DBLP:conf/ijcnn/WangLLZ19}  \cite{DBLP:conf/starsem/SergiehBGR18}. Each entity in these MKGs contains three modal information: structure, image, and text. Specifically, the relation triplets and textual descriptions of two datasets are extracted from WordNet \cite{DBLP:journals/cacm/Miller95} and Freebase \cite{DBLP:conf/sigmod/BollackerEPST08}. To extract the image features of the entities, 10 images and 100 images are crawled for each entity in WN9-IMG-TXT and FB-IMG-TXT, respectively \cite{DBLP:conf/starsem/SergiehBGR18}.  
Detailed statistics are shown in Table \uppercase \expandafter{\romannumeral2}.

\subsubsection{Evaluation Protocol} 
Entity link prediction \cite{DBLP:conf/nips/BordesUGWY13} and relation link prediction  \cite{DBLP:conf/sigir/NiuLTGDLWSHS21} are used to evaluate the performance of MMKGR. For \textbf{entity link prediction}, we use the following two metrics.  (1) Mean reciprocal rank (MRR) of all correct entities and (2) the proportion of correct entities that rank no larger than
N (Hits@N) \cite{DBLP:conf/emnlp/LvGHHLL19, DBLP:conf/emnlp/LinSX18}. For \textbf{relation link prediction}, mean average precision (MAP) score is adopted as the metric \cite{DBLP:conf/ijcai/WanP00H20}.

\subsubsection{Hyper-parameters}In the training step, key hyper-parameter settings are as follows. The embedding dimension \emph{$d_s$} of entity, relation and history is set to 200, the embedding dimension \emph{$d_i$} of image feature is set to 128 and 4096 on FB-IMG-TXT and WN9-IMG-TXT respectively, and the embedding dimension \emph{$d_t$} of textual feature is 1000 \cite{DBLP:conf/starsem/SergiehBGR18}. The maximum reasoning step \emph{T} is set to 4. The size of batches $\emph{N}$ is 128. The bandwidth $u$ in Eq. (15) is set to 3. $\lambda_{1}$, $\lambda_{2}$, and $\lambda_{3}$ in Eq. (16) are set to 0.1, 0.8, and 0.1, respectively. 

\subsection{Baselines}
To investigate the performance of MMKGR, two categories of methods are compared: 1) single-hop MKG reasoning methods MTRL \cite{DBLP:conf/starsem/SergiehBGR18}; 2) multi-hop reasoning methods MINERVA \cite{DBLP:conf/iclr/DasDZVDKSM18}, FIRE \cite{DBLP:conf/emnlp/Zhang0S0C20}, GAATs \cite{DBLP:journals/access/RocktaschelR17}, NeuraILP \cite{DBLP:conf/nips/YangYC17} and RLH  \cite{DBLP:conf/ijcai/WanP00H20} in traditional KGs. 

\subsection{Performance Comparisons}

Link prediction results are illustrated in Table \uppercase \expandafter{\romannumeral3} (all scores are in percentage) and \uppercase \expandafter{\romannumeral4}, where the results of most competitive baselines are marked by underline and the best results are highlighted in bold. We have the following observations.

\subsubsection{Entity Link Prediction}
We study entity link prediction from Table \uppercase \expandafter{\romannumeral3}. Firstly, the scores on FB-IMG-TXT are generally lower than those on WN9-IMG-TXT as shown in Table \uppercase \expandafter{\romannumeral3}, which is consistent with previous work. The essential reason is that the dataset FB-IMG-TXT is more sparse and complex than the dataset WN9-IMG-TXT \cite{DBLP:conf/starsem/SergiehBGR18} as shown in Table \uppercase \expandafter{\romannumeral2}. In addition, Hits@1 has more improvement than Hits@10 or Hits@5, which indicates that 
MMKGR tends to rank the ground-truth entity higher and has superior reasoning ability than other models.

Secondly, although multi-modal features are not used, some RL-based methods (e.g., RLH) and the graph neural network-based  reasoning method (i.e., GAATs) exceed MTRL that uses multi-modal features in overall performance. There are two potential reasons. On the one hand, the performance of MTRL is limited by TransE-based single-hop model. On the other hand, these novel models make better use of structural data (e.g.,  neighbor entities) in the multi-hop reasoning process.

Finally, our model achieves the best performance among all the methods. 
Observed from the last row of Table \uppercase \expandafter{\romannumeral3}, the
most significant improvements on the two datasets are 17.8\% and
21.3\%, respectively. In general, the above results prove that our model significantly outperforms existing methods in most metrics.

\begin{table*}
	\centering
	\caption{Effects of Different Multi-modal Auxiliary Features on Entity Link Prediction.}
	\setlength{\tabcolsep}{3.3mm}{
		\begin{tabular}{c|cccc|cccc}
			\toprule
			&  \multicolumn{4}{c|}{WN9-IMG-TXT} & \multicolumn{4}{c}{FB-IMG-TXT} \\ 
			Model         & MRR    & Hits@1 & Hits@5    &Hits@10    & MRR        & Hits@1   & Hits@5    & Hits@10    \\ \midrule

			OSKGR     & 66.0     & 61.5 & 82.5  & 90.5& 55.1  & 47.8 & 63.1 & 73.2 \\
			STKGR     & 71.2     & 65.1 & 84.6  & 91.3 & 60.1  & 52.3 & 64.9  & 75.3 \\
			SIKGR     &{\underline{74.7}}     &{\underline{68.8}}    & {\underline{85.8}} &{\underline{91.9}}  &{\underline{66.8}}   &{\underline{59.7}}  & {\underline{69.4}}  &{\underline{78.6}} \\
			MMKGR    & \textbf{80.2}     & \textbf{73.6} &  \textbf{87.8} & \textbf{92.8} & \textbf{71.3}  & \textbf{65.8}  & \textbf{77.5} & \textbf{82.6} \\
			\bottomrule

		\end{tabular}
	}
	\label{tab:whoiswho}
	\vspace{-0.4cm}
\end{table*}

\subsubsection{Relation Link Prediction}

The result of relation link prediction is presented 
in Table \uppercase \expandafter{\romannumeral4}. The upper and lower parts of the table are the MAP scores on WN9-IMG-TXT and FB-IMG-TXT, respectively.  We compare the MAP score of each relation (e.g., has\_part and place\_founded), and then count the overall MAP score (i.e., ``Overall") of all relations in the test set. Observed from Table \uppercase \expandafter{\romannumeral4}, the MAP scores of different models on WN9-IMG-TXT are higher than those on the FB-IMG-TXT, which is similar to the distribution of the experimental results of entity prediction. In additon, the overall improvements of MMKGR compared with RLH on the
two datasets are 13.7\% and 19.0\%, respectively.
Regardless of the comparison over each relation or overall relations, the performance of MMKGR  surpasses other models. 

Based on the experimental results of entity and relation link prediction, we can summarize that MMKGR has made great progress in the multi-modal knowledge graph reasoning. This is because MMKGR makes full use of all modal data through the unified gate-attention network and effectively leverages these data to help multi-hop reasoning without being negatively affected by the sparse reward dilemma.

\begin{figure}
	\centering
	\subfigure[WN9-IMG-TXT]
	{
		\centering
		\includegraphics[width=0.466\linewidth,height=3.5cm]{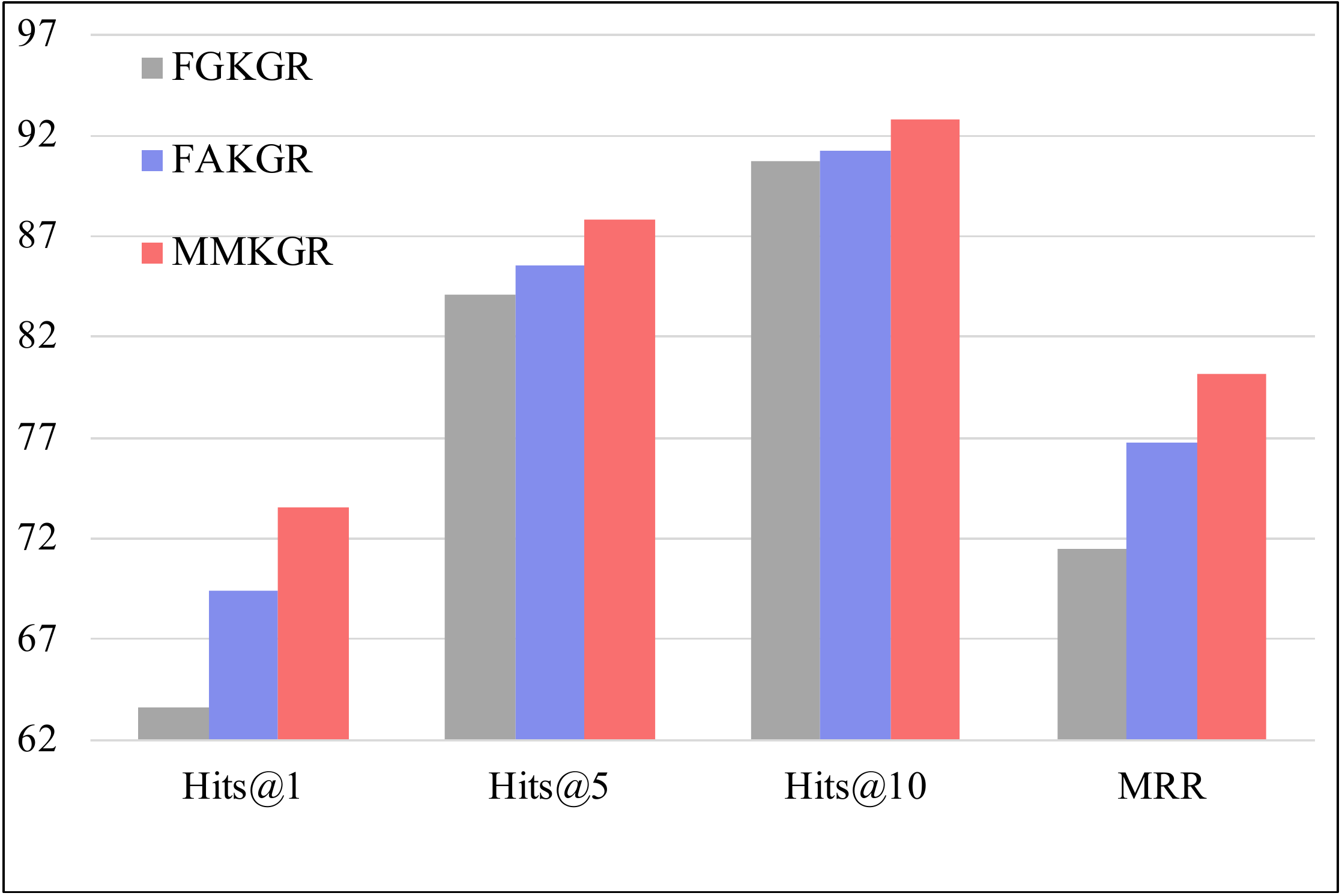}
	}
	\subfigure[FB-IMG-TXT]
	{
		\centering
		\includegraphics[width=0.466\linewidth,height=3.5cm]{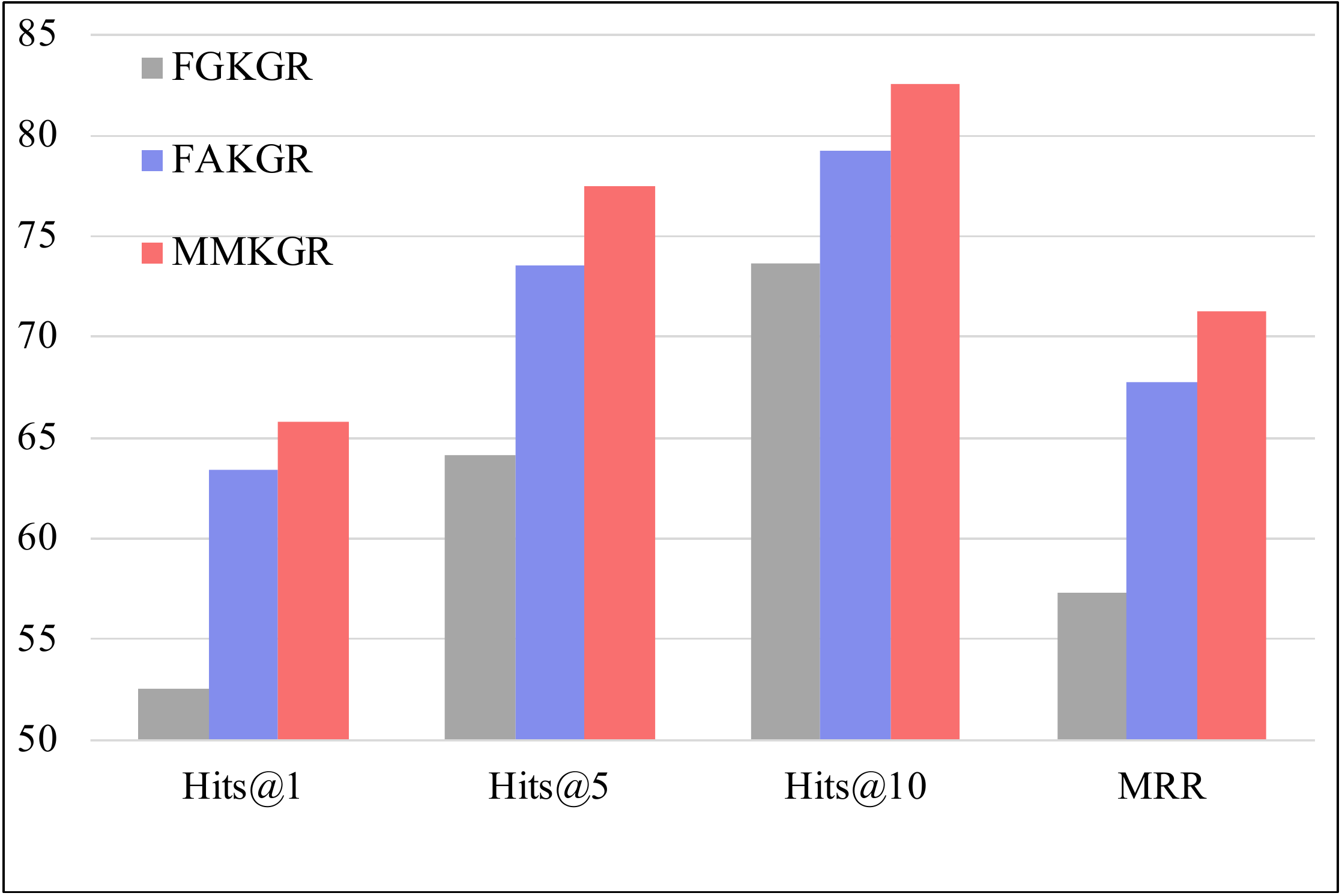}
	}
	\caption{Ablation on different components of the unified gate-attention network.}
	\label{fig:ablation}
	\vspace{-0.5cm}
\end{figure}

\subsection{Ablation Studies}
\subsubsection{Impact of  Different Components in the Unified Gate-attention Network}
We perform the ablation study by separately removing different components. 
(1)  \textbf{FAKGR}: the irrelevance-filtration module is removed in this variant version. The attended features $\hat{V}$ generated by the attention-fusion module are directly fed into the feature-aware RL framework. 
(2) \textbf{FGKGR}: after multi-modal fusions are completed by Eq. (6), only the irrelevance-filtration module is used to generate features that are fed into the complementary feature-aware RL. This variant version is to evaluate the effectiveness of the well designed attention-fusion module. 
The results are shown in Fig. 4. Several observations are made from the results. First, MMKGR has the best performance compared with both variant versions. This validates the effectiveness of both the attention-fusion module and the irrelevance-filtration module. Second, the results of FAKGR are closer to those of MMKGR and consistently better than those of FGKGR. The above results indicate that the main improvement of MMKGR comes from the attention-fusion module, and the improvement of the irrelevance-filtration module is relatively small. 
This is because WN9-IMG-TXT and FB-IMG-TXT have eliminated part of the influence of noise when the multi-modal data are crawled from Web \cite{DBLP:conf/starsem/SergiehBGR18}.  Third, even though FGKGR performs worse compared with the variant versions of our own models FAKGR and MMKGR, combined with the results in Table \uppercase \expandafter{\romannumeral3}, we can observe that the results of FGKGR are better than those of MTRL and RLH (i.e., the state-of-the-art baseline among MKG reasoning and KG reasoning methods).
For example, the Hits@1 of FGKGR is about 27.4\% and 8.1\% higher than that of MTRL and HRL, respectively. This demonstrates that the irrelevance-filtration module makes full use of multi-modal features and positively affects reasoning performance in MKGs.

\begin{figure}
	\centering
	\subfigure[WN9-IMG-TXT]
	{
		\centering
		\includegraphics[width=0.466\linewidth,height=3.5cm]{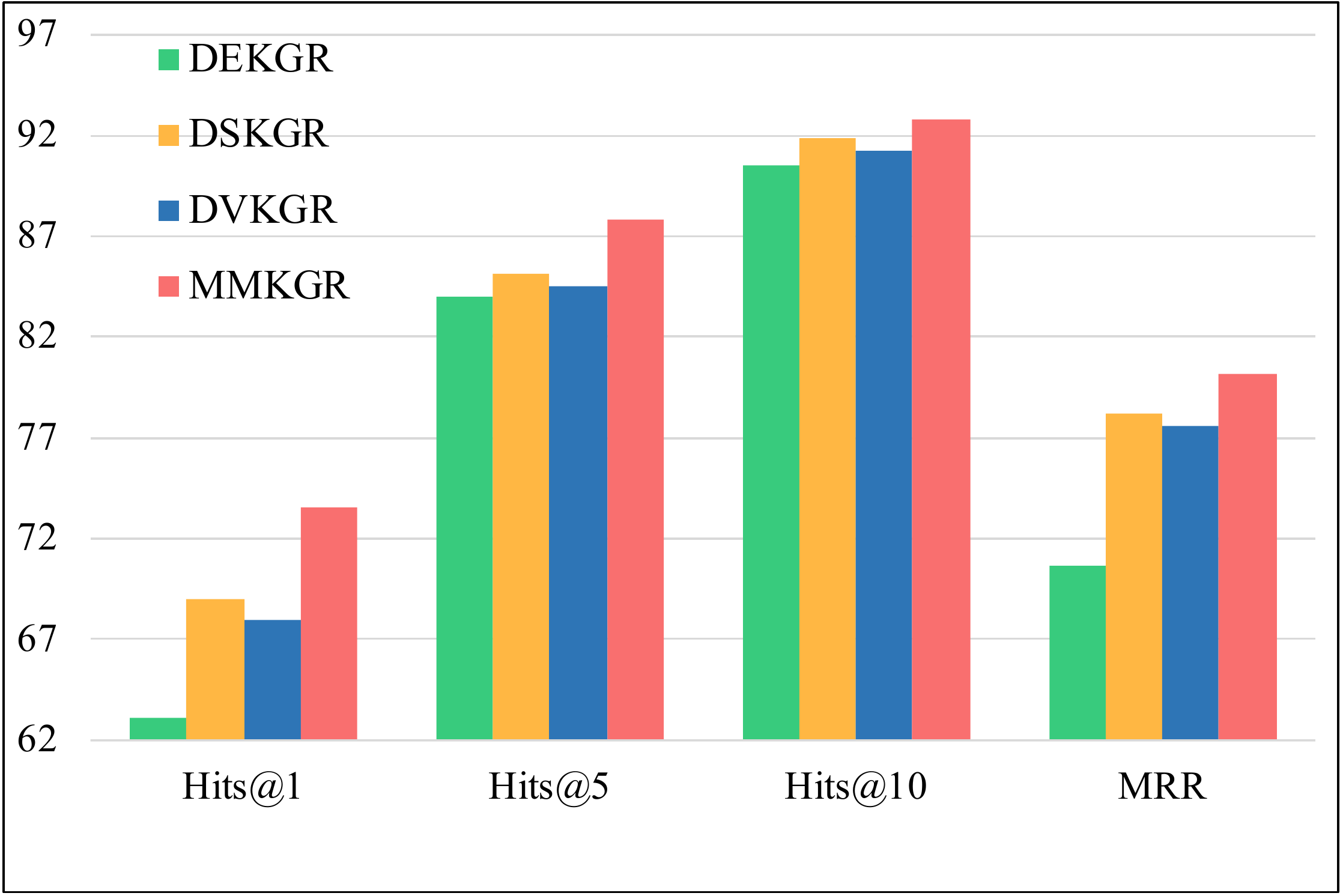}
	}
	\subfigure[FB-IMG-TXT]
	{
		\centering
		\includegraphics[width=0.466\linewidth,height=3.5cm]{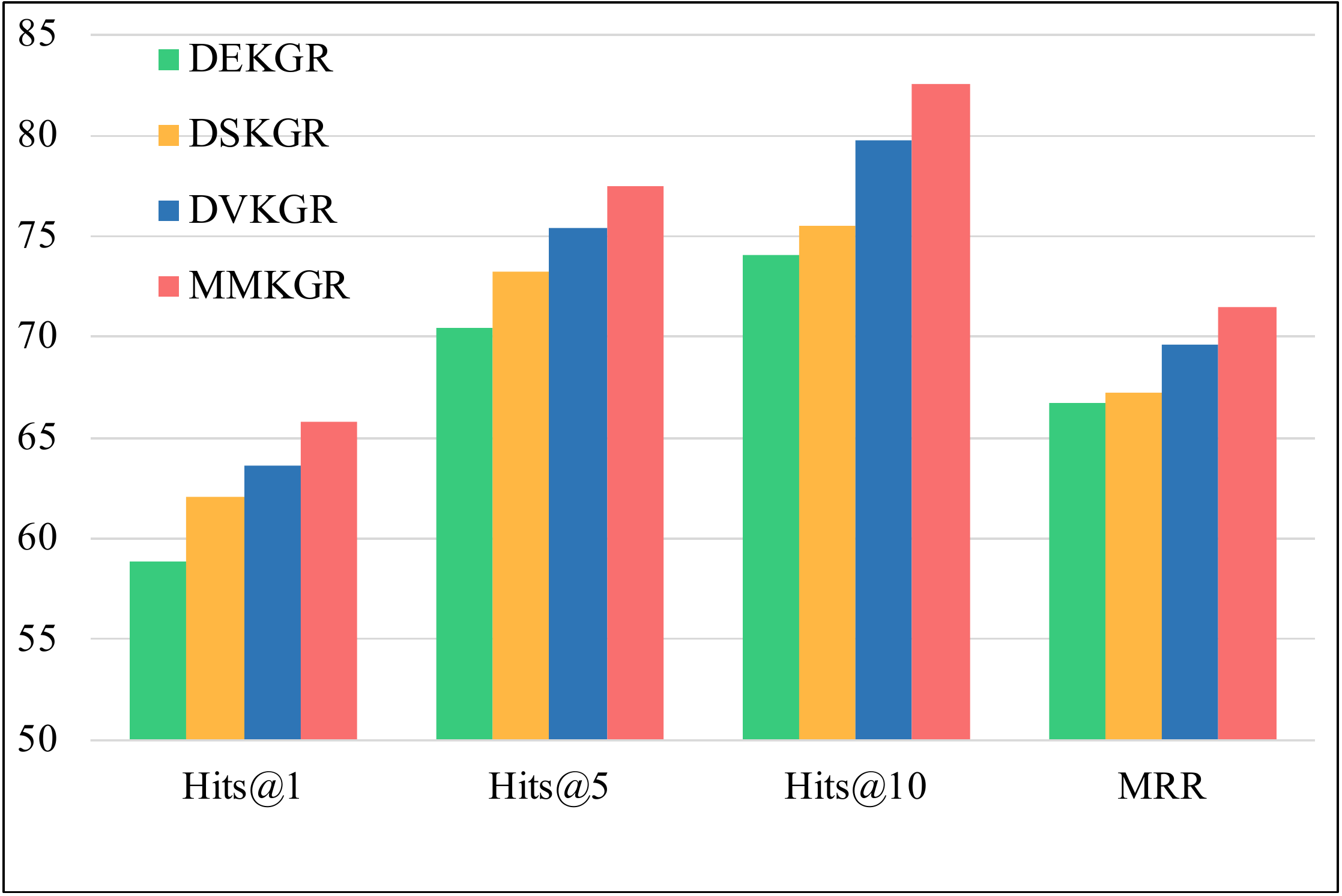}
	}
	\caption{Ablation on different components of 3D reward mechanism.}
	\label{fig:ablation}
	\vspace{-0.6cm}
\end{figure}

\begin{table*}
	\centering
	\caption{Hits@1 of MMKGR with the Changing Reasoning Step \emph{T} and Different Reward Threshold of  \emph{k}.}
	\setlength{\tabcolsep}{3.3mm}{
		\begin{tabular}{c|ccccc|ccccc}
			\toprule
			&  \multicolumn{5}{c|}{WN9-IMG-TXT} & \multicolumn{5}{c}{FB-IMG-TXT} \\ 
			\diagbox{Th.}{Hits@1}{\emph{T}}      & \emph{T}=2    & \emph{T}=3 & \emph{T}=4   & \emph{T}=5 &\emph{T}=6    & \emph{T}=2    & \emph{T}=3 & \emph{T}=4    &\emph{T}=5 &\emph{T}=6   \\ \midrule

			2    & 45.7     & 69.8 & 71.8  & 67.4& 64.8 & 47.9 & 60.5 & 62.8& 57.8 & 55.1 \\
			\textbf{3}     & ---     & \textbf{73.1} & \textbf{73.6}  & \textbf{73.5} & \textbf{73.3}  & --- & \textbf{65.3}  & \textbf{65.8} & \textbf{64.9} & \textbf{64.1}\\
			4    & ---     & --- & 72.1  & 71.5& 71.1  & --- & --- & 63.3& 62.4 & 61.6 \\
			5    & ---      & ---  &  ---  & 71.4 & 70.8  & ---  & ---  & ---  & 61.7 & 61.1\\
						6    & ---      & ---  &  ---  & --- & 70.7 & ---  & ---  & ---  & --- & 60.7\\
			\bottomrule

		\end{tabular}
	}
	\label{tab:whoiswho}
	\vspace{-0.5cm}
\end{table*}

\subsubsection{Impact of Different Components in the Reward Function}
The 3D reward mechanism can solve the sparse reward problem. It includes three components: the destination reward, the distance reward, and the diverse reward. Note that. the destination reward is indispensable reward setting used to drive the reasoning agent to the target entity. Based on this, we perform the ablation study for the reward function by separately removing different components of the 3D reward mechanism: (1) \textbf{DEKGR}, a variant version only leveraging the destination reward as the reward function; (2) \textbf{DSKGR}, a variant version where the distance reward is added on the basis of the destination reward; (3) \textbf{DVKGR}, in which diverse reward is added on the basis of the destination reward. 

In Fig. 5, we first observe that all three variant versions perform worse than MMKGR which demonstrates that
both diverse reward and distance reward can improve reasoning performance. Second, the reasoning performance fluctuates when any reward is removed. Furthermore, although DEKGR has the lowest performance among the variants, it still outperforms the state-of-the-art methods MTRL and RLH, which proves the effectiveness of destination reward in MKG reasoning. Finally, the performance of DSKGR and DVKGR is closer to that of MMKGR than the other variants on WN9-IMG-TXT and FB-IMG-TXT, respectively. This experimental result illustrates that more diverse paths need to be explored by diverse reward in larger dataset (i.e., FB-IMG-TXT), while lightweight datasets (i.e., WN9-IMG-TXT) rely more on the distance reward. A potential reason is that there are not enough diverse paths to be inferred in lightweight datasets.

\subsection{Effects of  Multi-modal Features}

In this subsection, we focus on the effect of different multi-modal auxiliary features over the reasoning result. To evaluate this, we compare three versions where features of a type of modality are removed: (1) \textbf{OSKGR}: a version where only structural features are considered in Eq. (17); (2) \textbf{STKGR}: a version where image features are not calculated by the unified gate-attention network. 
(3) \textbf{SIKGR}: a version that textual features are not input into the unified gate-attention network. 
Observed from Table \uppercase \expandafter{\romannumeral5}, OSKGR still outperforms existing multi-hop reasoning methods in Table \uppercase \expandafter{\romannumeral3}, even though the multi-modal features are not added. For example, the MRR of OSKGR is 3.6\% and 4.5\% higher than that of RLH on both datasets, respectively. This is because OSKGR retaining 3D reward mechanism can eliminate the negative impact of sparse rewards and improve reasoning   performance. In addition, the performance of MMKGR is significantly better than that of the other three variant versions which validates the benefit brought by both image and text features and confirms the necessity for including multi-modal data in KG reasoning.
Finally, the performance of SIKGR is better than that of STKGR. This is because each entity is connected to a larger number of pictures (i.e., 100 images for each entity on FB-IMG-TXT) containing more useful information.

\renewcommand{\arraystretch}{1} 
\begin{table}
\tabcolsep 3.6pt
  \centering
  \begin{threeparttable}
  \caption{
The performance change of models after fusion of multi-modal information on FB-IMG-TXT.}
  \label{tab:performance_comparison}
    \begin{tabular}{ccccccc}
    \toprule
    \multirow{2}{*}{Method}&
    \multicolumn{2}{c}{Attention}&\multicolumn{2}{c}{Concatenation}\cr
    \cmidrule(lr){2-3} \cmidrule(lr){4-5} 
    &Rewards&Hits@1&Rewards&Hits@1\cr
    \midrule
    GAATs &--- &-2.1\%  &--- &-3.7\% \cr
    NeuralLP &--- &-3.3\%  &--- &-5.4\% \cr
    MINERVA&-2.2\%&-6.3\% &-2.7\% &-7.1\%  \cr
    FIRE&-1.8\% &-5.9\%  &-2.3\% &-6.5\% \cr
    RLH &-1.1\% &-3.8\%  &-1.7\% &-4.9\% \cr
    \bottomrule
    \end{tabular}
    \end{threeparttable}
\vspace{-0.45cm}
\end{table}

\begin{table}

  \centering
 
  \begin{threeparttable}
  \caption{Comparison of Hits@1 on test sets with different proportions.}
  \label{tab:performance_comparison}
    \begin{tabular}{ccccccc}
    \toprule
    \multirow{2}{*}{Proportion}&
    \multicolumn{2}{c}{WN9-IMG-TXT}&\multicolumn{2}{c}{FB-IMG-TXT}\cr
    \cmidrule(lr){2-3} \cmidrule(lr){4-5}
    & MMKGR        & OSKGR  & MMKGR  & OSKGR\cr
    \midrule
    20\% &   \textbf{85.6}       &    74.1 &  \textbf{60.8} &  40.2 \cr
    40\% &    \textbf{75.5}      &  65.0  & \textbf{71.8}  &   59.3 \cr
    60\%&  \textbf{72.3}      &   60.4    &  \textbf{68.7}       &   54.9 \cr
    80\%&\textbf{69.4}      &  60.1     &  \textbf{57.6}       &  41.1  \cr
    100\% & \textbf{73.6}     &  61.5    &  \textbf{65.8} &  47.8  \cr
    \bottomrule
    \end{tabular}
    \end{threeparttable}
\vspace{-0.5cm}
\end{table}

\begin{figure*}
	\centering
	\subfigure[MMKGR]
	{
		\centering
		\includegraphics[width=0.19\linewidth]{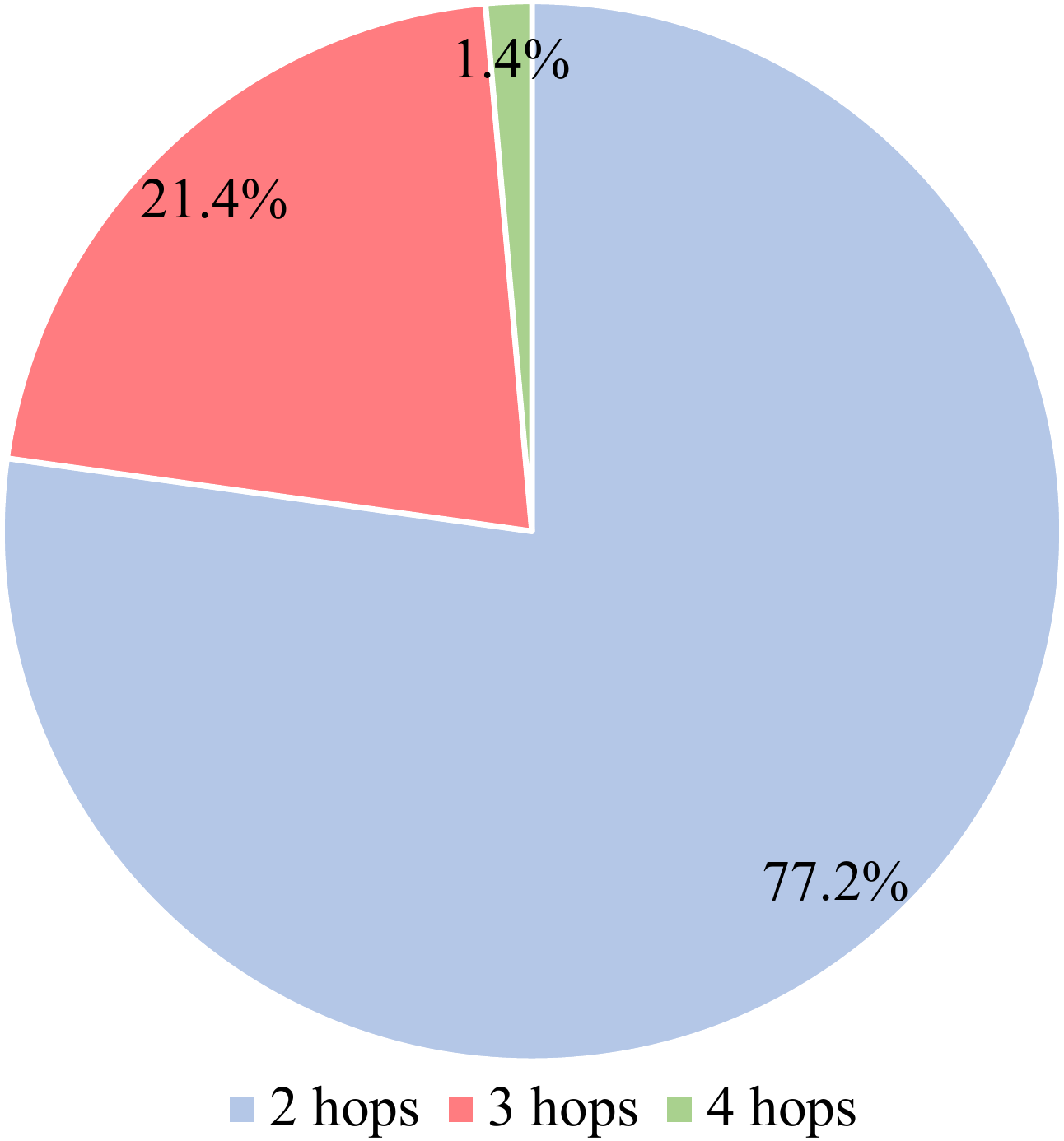}
	}
	\subfigure[DVKGR]
	{
		\centering
		\includegraphics[width=0.19\linewidth]{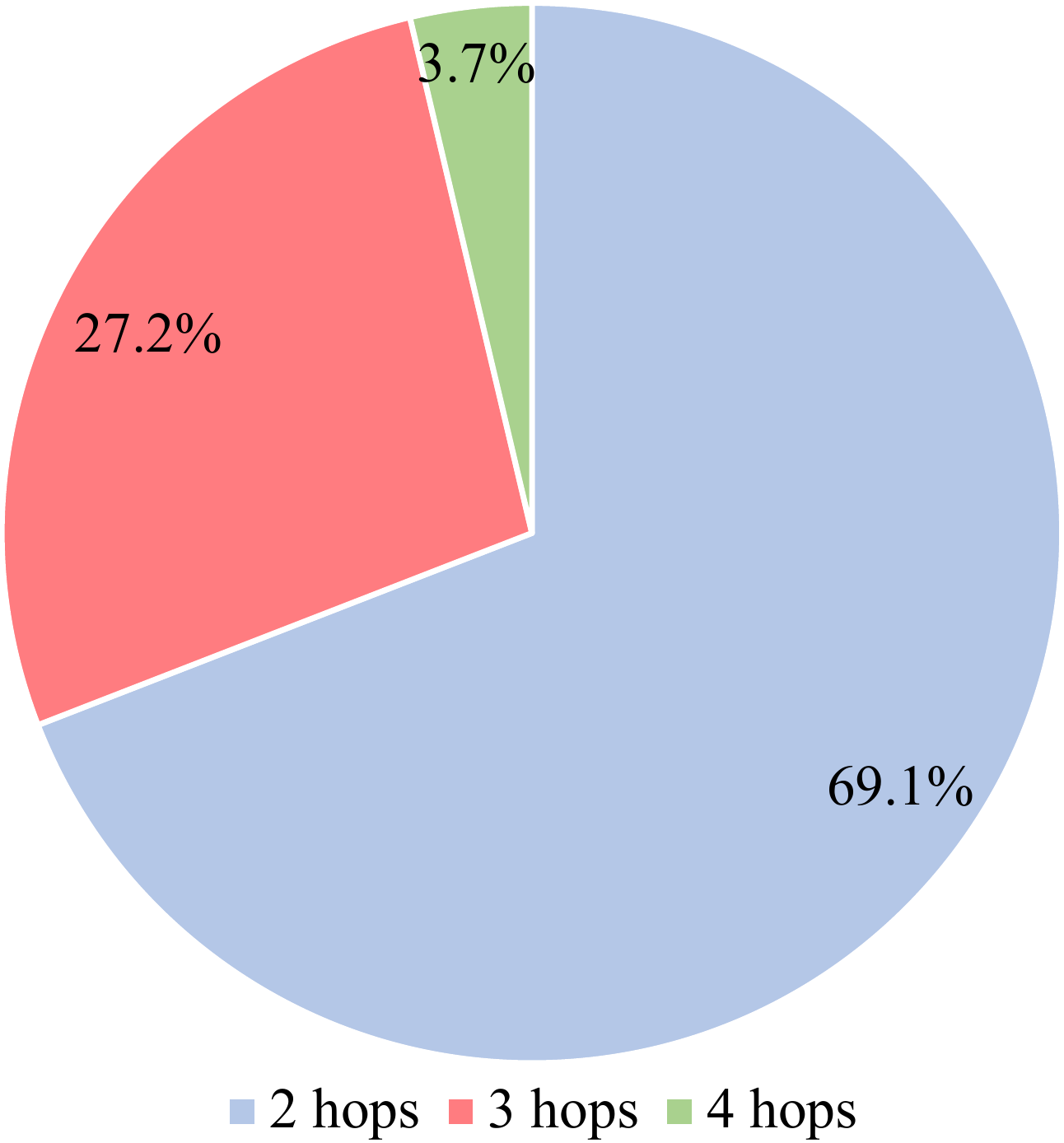}
	}
	\subfigure[OSKGR]
	{
		\centering
		\includegraphics[width=0.19\linewidth]{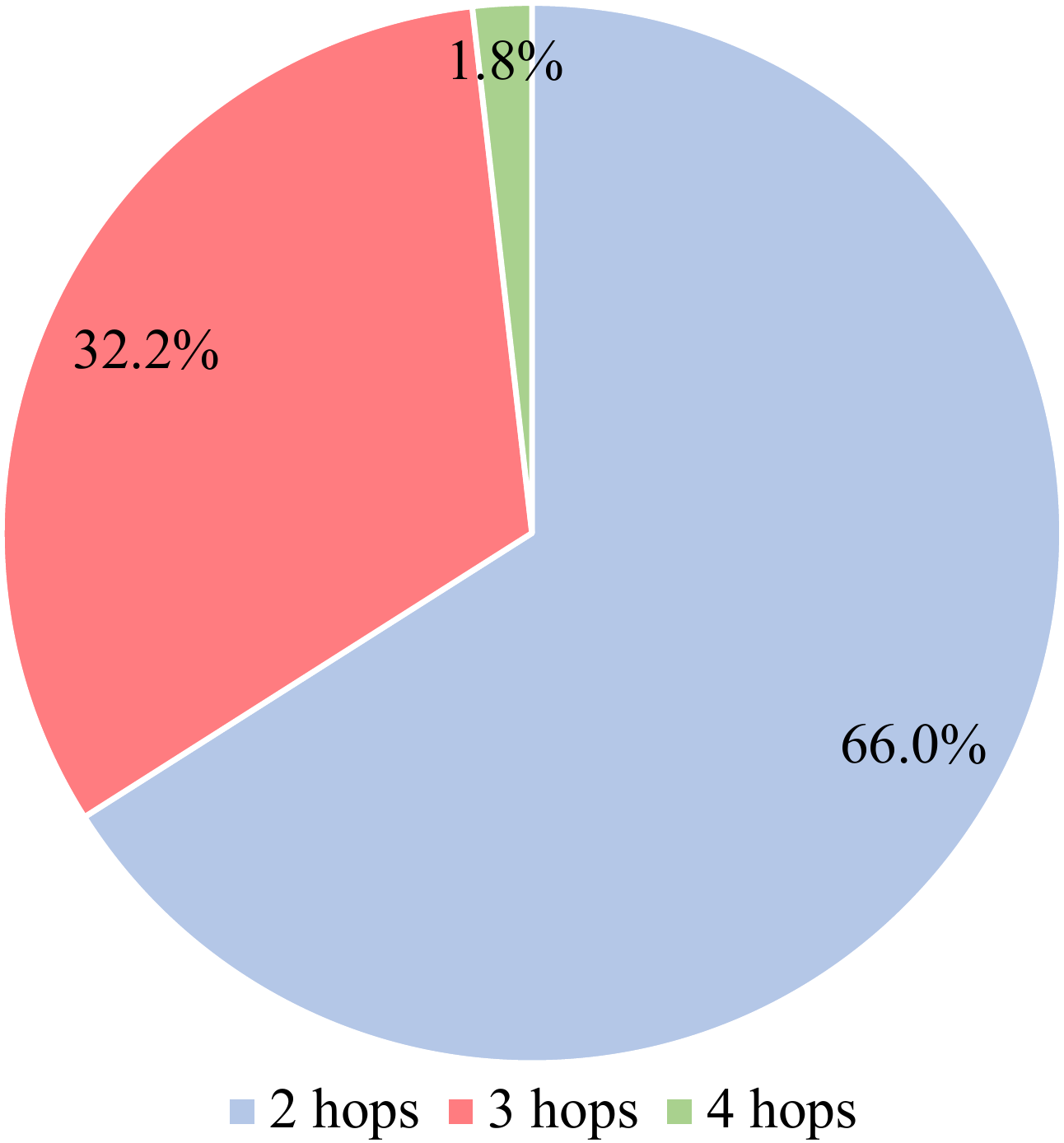}
	} 
	
	\caption{The proportions of test triplets successfully inferred by different path lengths on WN9-IMG-TXT.}
	\label{fig:ablation}

\end{figure*}

\begin{figure*}
	\centering
	\subfigure[MMKGR]
	{
		\centering
		\includegraphics[width=0.19\linewidth]{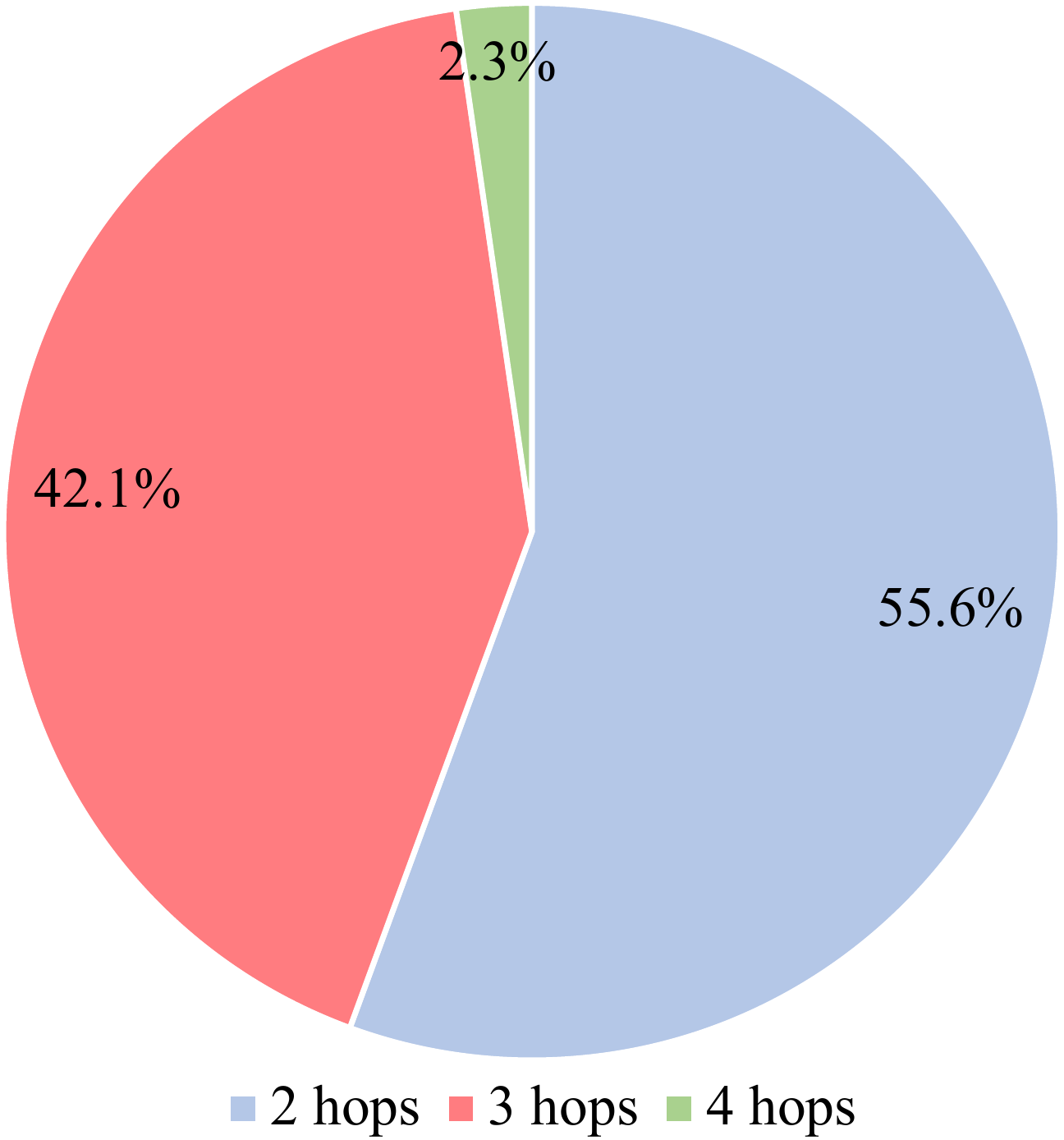}
	}
	\subfigure[DVKGR]
	{
		\centering
		\includegraphics[width=0.19\linewidth]{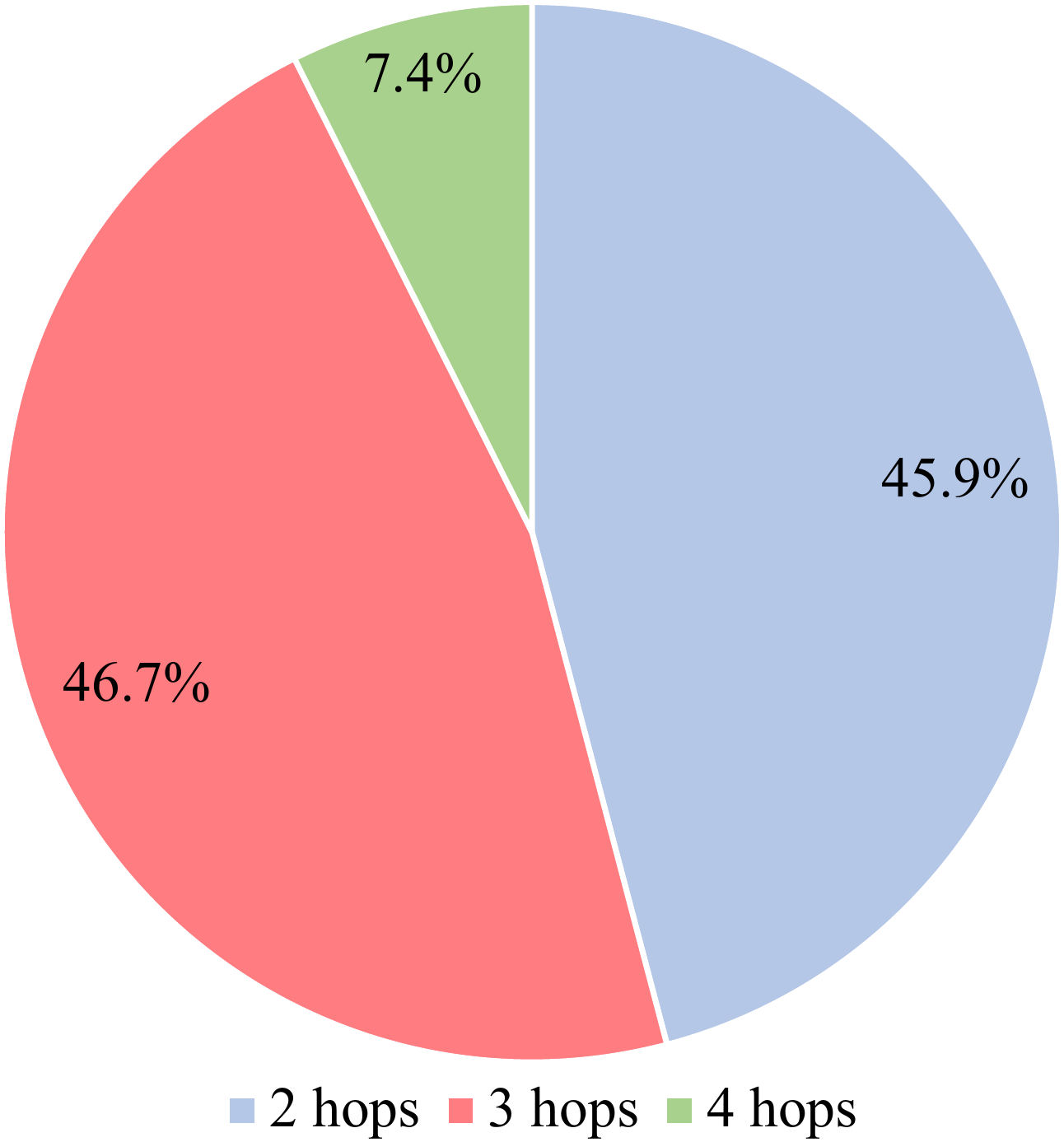}
	}
	\subfigure[OSKGR]
	{
		\centering
		\includegraphics[width=0.19\linewidth]{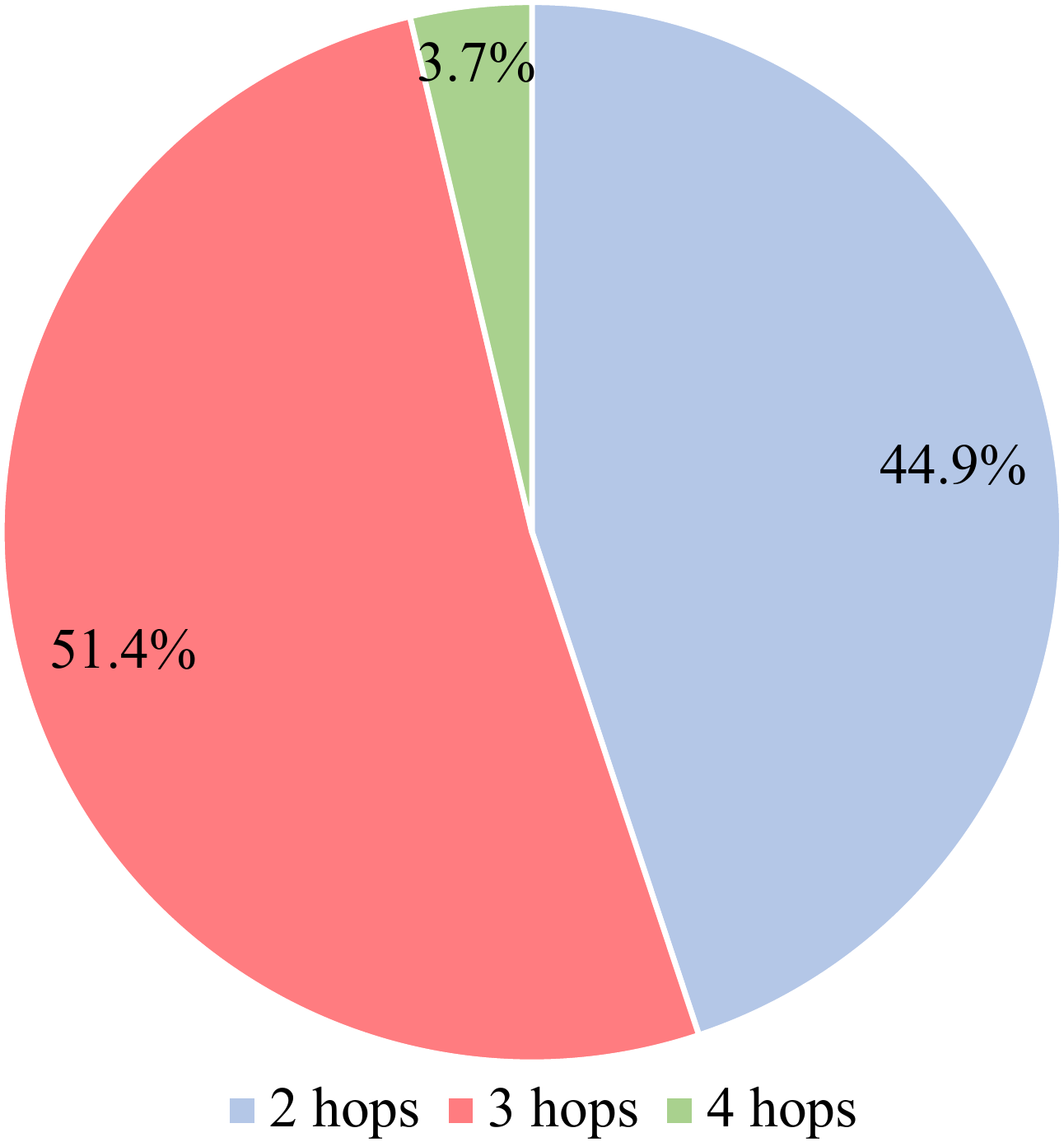}
	} 
	\caption{The proportions of test triplets successfully inferred  by different path lengths on FB-IMG-TXT.}
	\label{fig:ablation}
	\vspace{-0.3cm}
\end{figure*}

In fact, performing multi-hop reasoning in MKGs is challenging. To investigate the impact of this challenge on existing multi-hop reasoning models, we conduct the following experimental setup. Specifically, we first utilize  Attention  and Concatenation (i.e., two multi-modal fusion method derived from existing single-hop methods on MKGs). Then, existing multi-hop reasoning methods (e.g. GAATs) are we combined to conduct multi-hop reasoning on the MKG. 
As shown in Table \uppercase \expandafter{\romannumeral7}, when multi-modal data are added to the existing multi-hop reasoning models, all performance percentages of RL-based models decrease compared with the absence of these data. This is because sparse rewards reduce reward accumulation and limit reasoning performance on the MKG.  Although non-RL methods (e.g., GAATs) are not affected by the sparse rewards, the Hits@1 still declines compared with the absence of multi-modal data. A reasonable explanation is that existing fusion methods are not suitable for multi-hop reasoning methods.

To further verify the global impact of multi-modal features on multi-hop reasoning, we compare the changes of Hits@1 before and after fusing multi-modal data (i.e., feature fusion settings like OSKGR and MMKGR) on test sets. We first randomly sample different proportions of test data (e.g., 20\%), and then use MMKGR and OSKGR to perform reasoning. The experimental results are presented in Table \uppercase \expandafter{\romannumeral8}. We can observe that regardless of the proportion of test data, the Hits@1 of  MMKGR is much larger than that of OSKGR. This expected result once again proves that multi-modal auxiliary features can help the model to improve the reasoning performance. Furthermore, although it is challenging to integrate multi-modal auxiliary features with multi-hop reasoning, MMKGR is able to stably and effectively utilize these  features.

\begin{figure}
	\centering
	\subfigure[WN9-IMG-TXT]
	{
		\centering
		\includegraphics[width=0.466\linewidth,height=3.5cm]{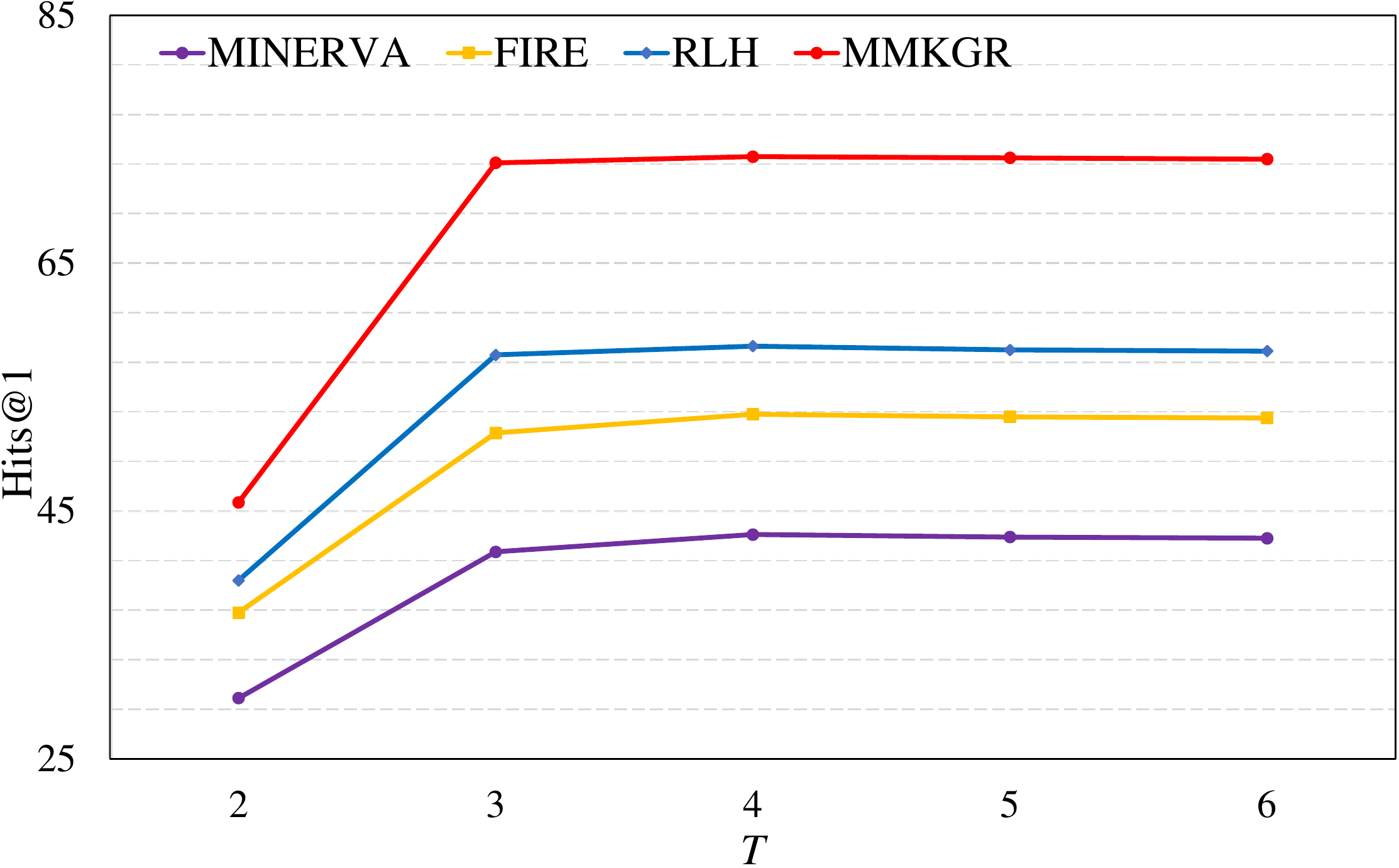}
	}
	\subfigure[FB-IMG-TXT]
	{
		\centering
		\includegraphics[width=0.466\linewidth,height=3.5cm]{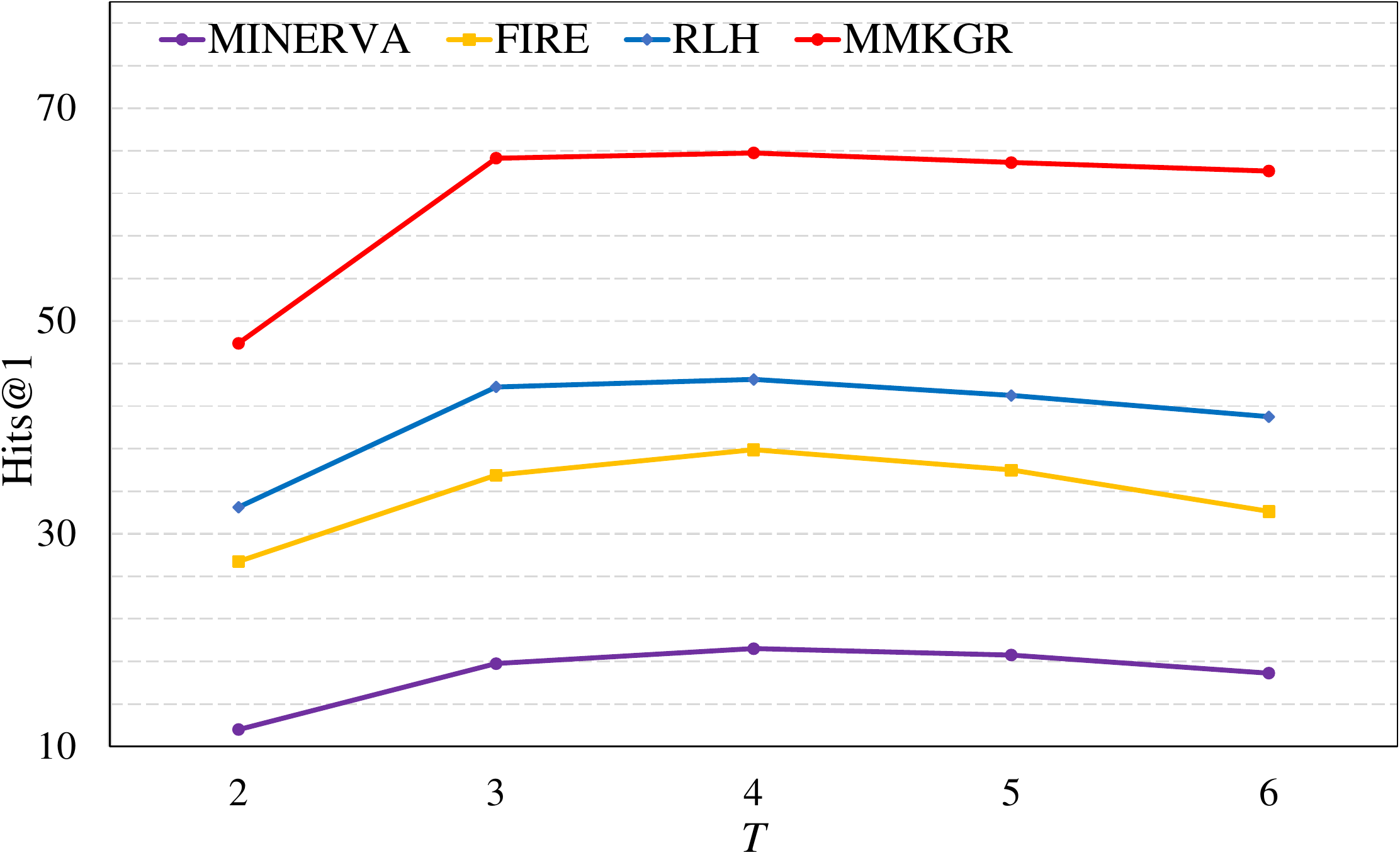}
	}
	\caption{Hits@1 of the changing reasoning step \emph{T} for RL-based models.}
	\label{fig:ablation}
	\vspace{-0.6cm}
\end{figure}
\subsection{Studies for Multi-hop Paths}
In this section, we study the factors that affect multi-hop paths. The first is the effect of distance reward in MMKGR. To investigate the plausibility that the threshold of \emph{k}  is set to 3 in the distance reward (i.e., Eq.(14)), we present corresponding experimental results in Table \uppercase \expandafter{\romannumeral6}. The horizontal line represents the null value because the threshold cannot be greater than the maximum reasoning step \emph{T}. Observed from  Table \uppercase \expandafter{\romannumeral6}, MMKGR achieves the best reasoning performance on both datasets when threshold of \emph{k} is set to 3. Note that when $k$ is the same as $T$, there is no negative penalty term in Eq. (14). In addition, Hits@1 has the fastest growth rate when the maximum reasoning step  \emph{T}=3 for all models as shown in Fig. 8.  Thus, we believe that long reasoning paths are also not required on MKGs.
After \emph{T}\textgreater4, the change of Hits@1 is relatively stable on the small-scale WN9-IMG-TXT, because most of the test triplets are successfully inferred before \emph{T}=4. Furthermore, the Hits@1 of all models gradually decreases when \emph{T}\textgreater4 on FB-IMG-TXT. One potential reason is that  some test triples negatively affected by noise are inferred incorrectly  after \emph{T}\textgreater4. Although the number of reasoning step exceeding the threshold 3 may degrade reasoning  performance as mentioned in Section \ref{Complementary Feature-aware Reinforcement Learning}, we can observe that reasoning performances of all models are still slightly improved when  \emph{T}=4. Therefore,  \emph{T} is set to 4 in this study.

As the first work for performing multi-hop reasoning in MKGs, it is necessary to investigate the affect of multi-modal data on the path length of MMKGR. For this purpose, we compare MMKGR with the ablated model OSKGR that only utilizes structural data. As another variable affecting reasoning length in this study, the role of the distance reward that encourages the agent to infer the target within 3 hops also needs to be investigated. Thus, we also compare MMKGR with the ablated model DVKGR that removes the distance reward from 3D reward in this subsection. The different proportions of triplets in test sets are inferred with different hops in Fig. 6 and 7, based on which we have the following two observations. (1) the proportion of 4 hops in DVKGR is the largest compared with MMKGR on both datasets. This shows that distance reward can further encourage the agent to find the target entity within the 3 hops most relevant to the query. (2) MMKGR has more 2-hop ratio and less 3-hop ratio than OSKGR. This indicates that multi-modal data provide beneficial reasoning clues to improve reasoning performance.

\subsection{Convergence Rate Analysis}

\begin{figure}
	\centering
	\subfigure[WN9-IMG-TXT]
	{
		\centering
		\includegraphics[width=0.45\linewidth,height=3.2cm]{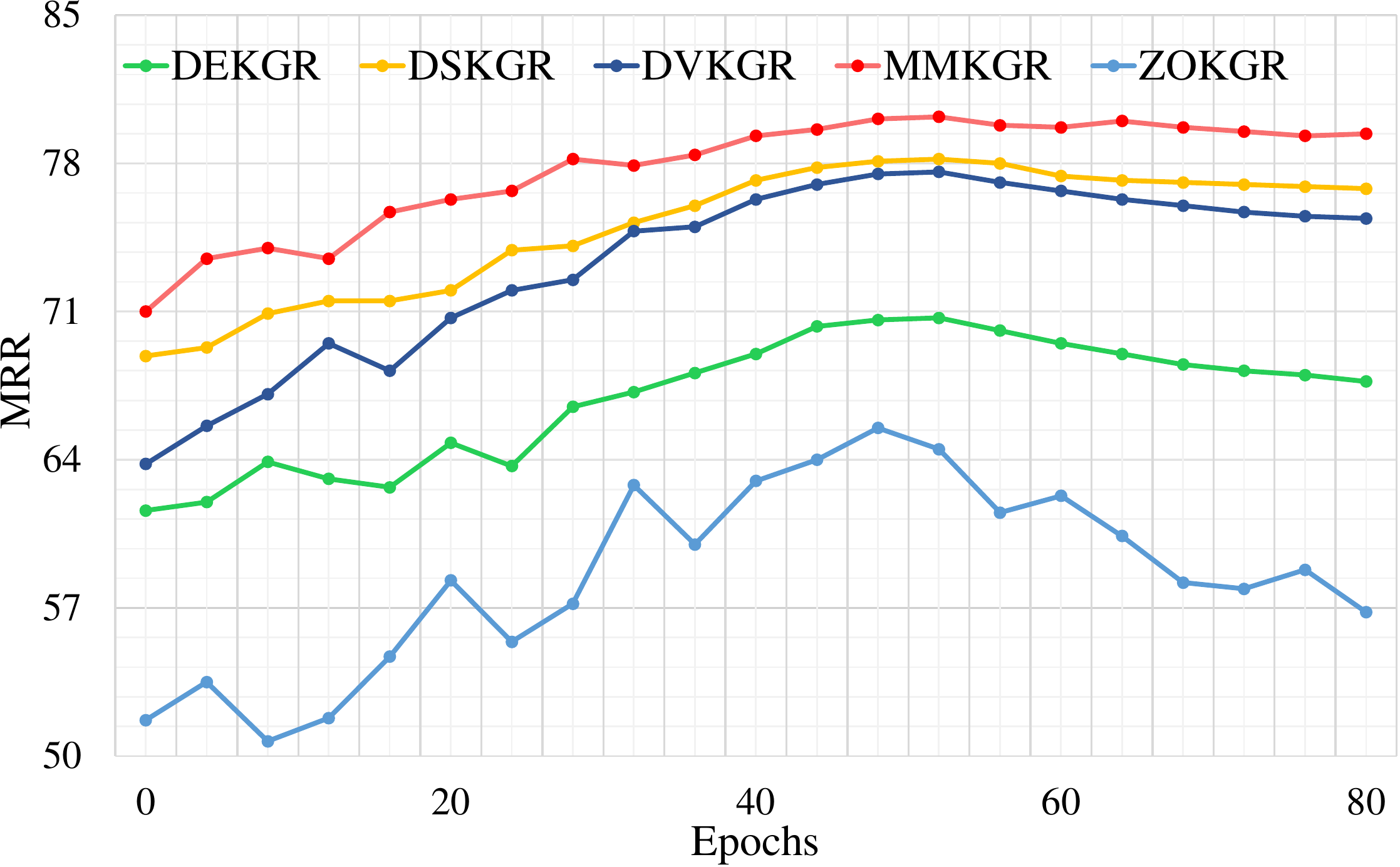}
	}
	\subfigure[FB-IMG-TXT]
	{
		\centering
		\includegraphics[width=0.45\linewidth,height=3.2cm]{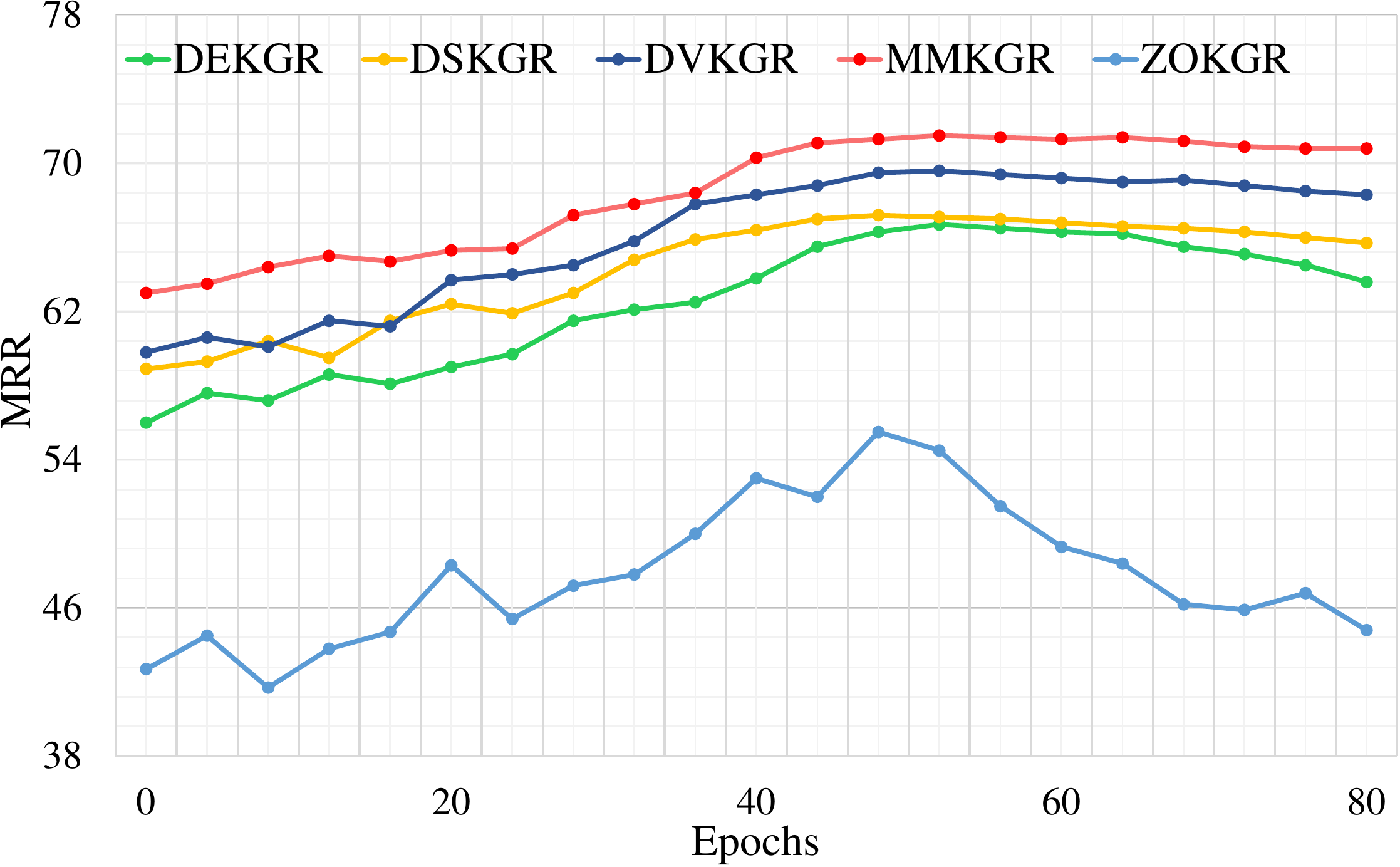}
	}
	\caption{The convergence rate of different methods.}
	\label{fig:ablation}
	\vspace{-0.6cm}
\end{figure}

\begin{figure}
	\centering
	\subfigure[WN9-IMG-TXT]
	{
		\centering
		\includegraphics[width=0.46\linewidth,height=3.2cm]{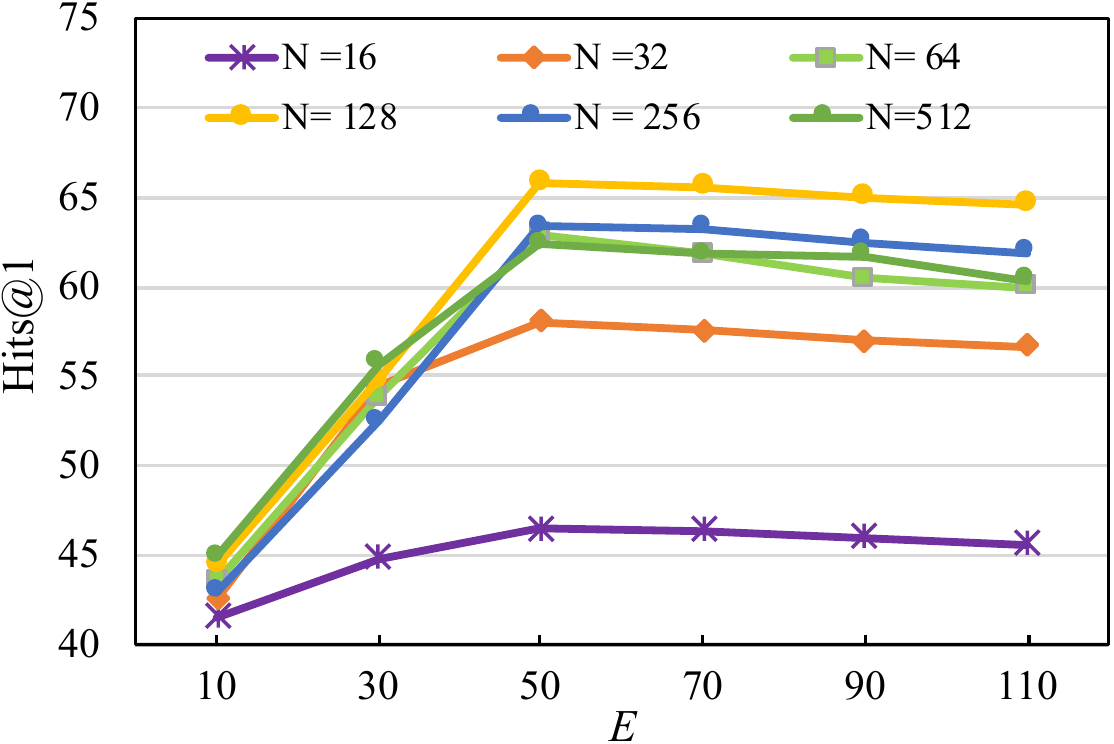}
	}
	\subfigure[FB-IMG-TXT]
	{
		\centering
		\includegraphics[width=0.46\linewidth,height=3.2cm]{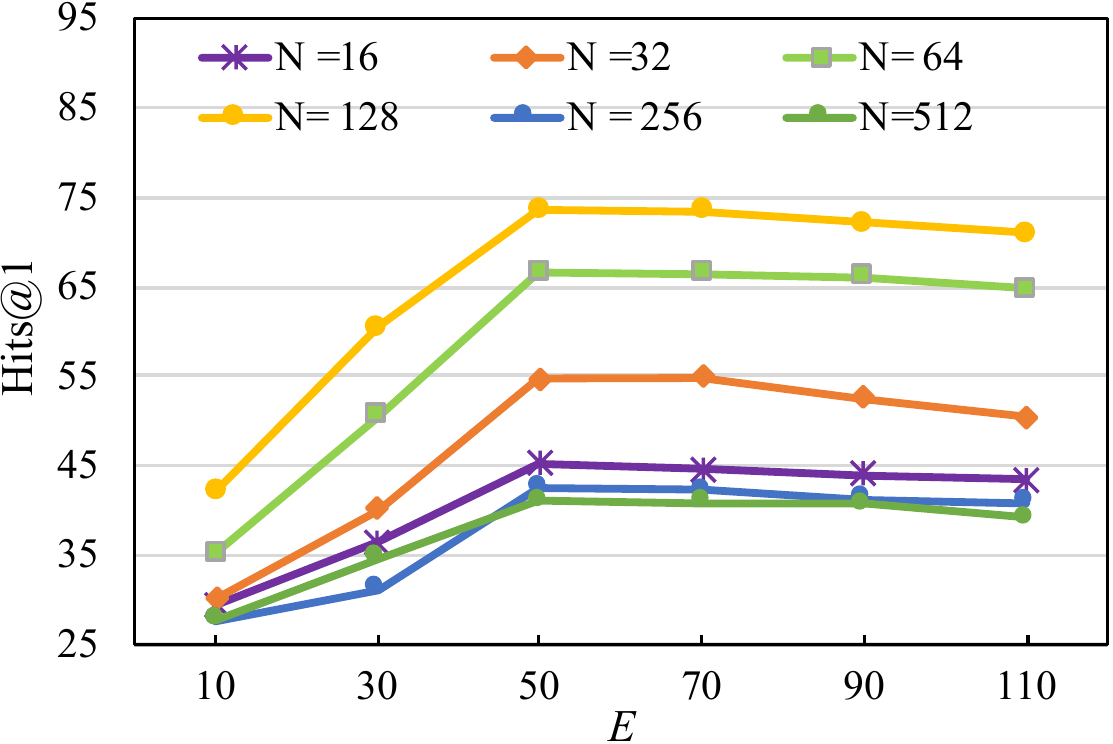}
	}
	\caption{The Hits@1 of MMKGR w.r.t. varied epoch $E$ and batch size $N$.}
	\label{fig:ablation}
\vspace{-0.3cm}
\end{figure}

We investigate the convergence rate of our proposed model in Fig. 9, where only the effect of 3D reward mechanism on convergence rate is presented, due to the space limitation. A new variant of MMKGR named ZOKGR is added. Notably, in ZOKGR, the 3D reward is completely replaced by the ``0-1 reward'' that is commonly used in existing RL-based reasoning methods (e.g. MINERVA and RLH). 

Observed from Fig. 9, ZOKGR fluctuates greatly and has not converged. A reliable  explanation is that the ``0-1 reward'' is easy to fall into the dilemma of sparse reward in MKGs. It is interesting that the MRR value of DEKGR on FB-IMG-TXT is closer to that of MMKGR. A potential reason is that our destination reward based on reward shaping plays a pivotal role in convergence rate on large datasets that are more likely to result in sparse rewards. In addition, a fair concern for diverse reward is that  it can slow down the convergence rate, since this reward encourages the agent to explore a more diverse set of paths. The good convergence of DVKGR eliminates the above concern. The convergence rates of DSKGR and DVKGR are both greater than that of DEKGR, which shows that both the distance reward and the diverse 
reward can accelerate the convergence.
In a word, the carefully designed 3D reward mechanism can boost reasoning performance and increase convergence rate on all datasets.

\subsection{Parameter Interpretability}

\subsubsection{Impact of Different Epochs $E$ and Batch Sizes $N$ }
We investigate the impact of different epochs $E$ and batch sizes $N$ in Fig. 10. Here, we set the number of epochs $E$ $\in$ [10, 30, 50, 70, 90, 110] and the size of batches $N$ $\in$ [16, 32, 64, 128, 256, 512]. We can observe that the performance of MMKGR will increase firstly and then decreases steadily in most cases with the increase of $E$ and $N$. This is because the under-training and over-fitting have a negative impact on the model. The results suggest that a proper size of train parameters can enhance the effectiveness of reasoning models. Consequently, the best parameters are chosen as $E$ = 50 and $N$ = 128.

\subsubsection{Impact of Different Bandwidths $u$}

\begin{figure}
	\centering
	\subfigure[WN9-IMG-TXT]
	{
		\centering
		\includegraphics[width=0.46\linewidth,height=3.3cm]{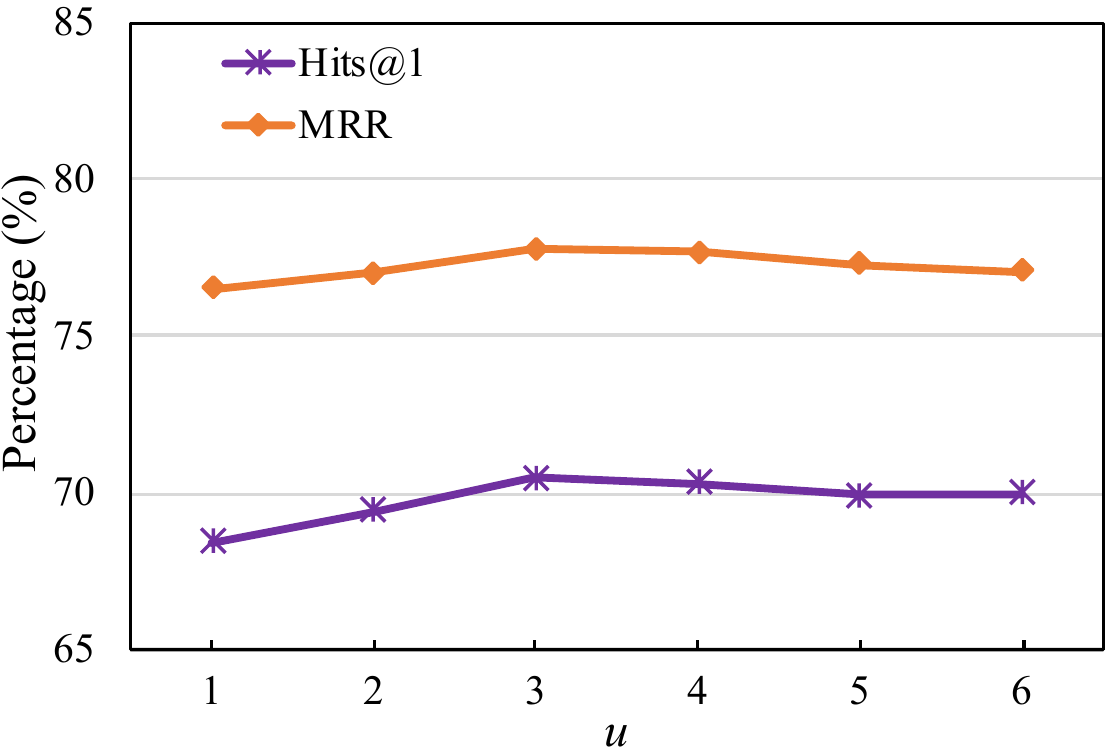}
	}
	\subfigure[FB-IMG-TXT]
	{
		\centering
		\includegraphics[width=0.46\linewidth,height=3.3cm]{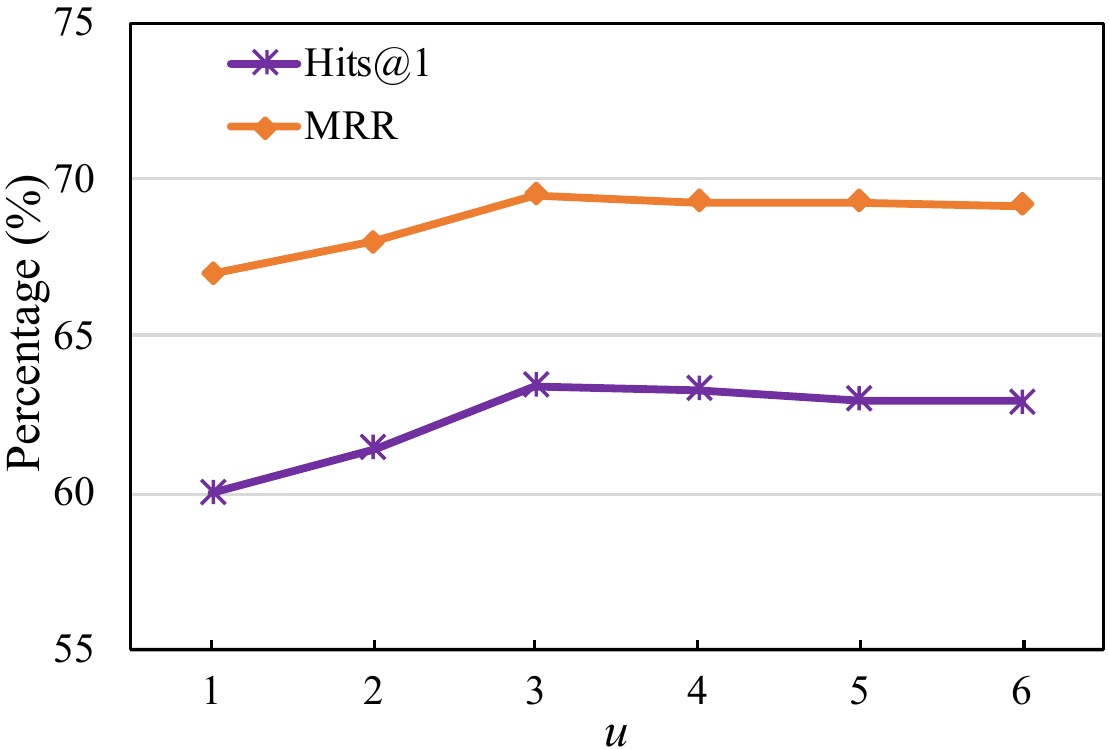}
	}
	\caption{Performance of MMKGR w.r.t. varied $u$ on different datasets.}
	\label{fig:ablation}
	\vspace{-0.35cm}
\end{figure}

We evaluate the influence of bandwidth $u$ on MMKGR. From the results in Fig. 11, we can observe that 3 is the optimal value of $u$ on both two datasets. If the value of $u$ exceeds 3, the performance will be relatively stable. 
One potential reason is that the  range of the local influence of the Gaussian kernel function increases as the bandwidth value increases. Beyond this range, the value of the kernel function is almost unchanged, and the impact on the reward becomes stable.

\subsubsection{Impact of Different Discount Factors $\lambda$}

\begin{figure}
	\centering
	\subfigure[WN9-IMG-TXT]
	{
		\centering
		\includegraphics[width=0.466\linewidth,height=3.5cm]{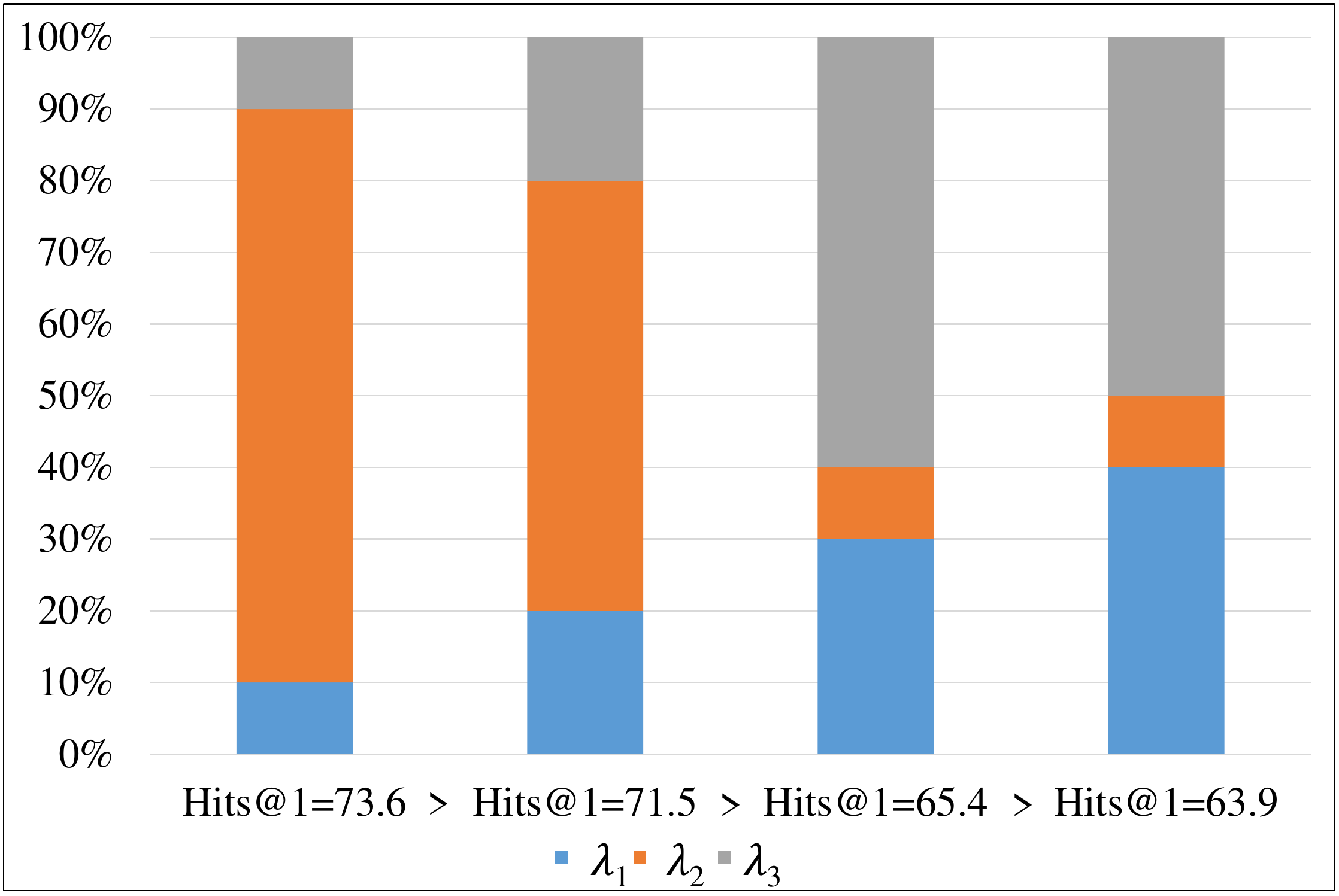}
	}
	\subfigure[FB-IMG-TXT]
	{
		\centering
		\includegraphics[width=0.466\linewidth,height=3.5cm]{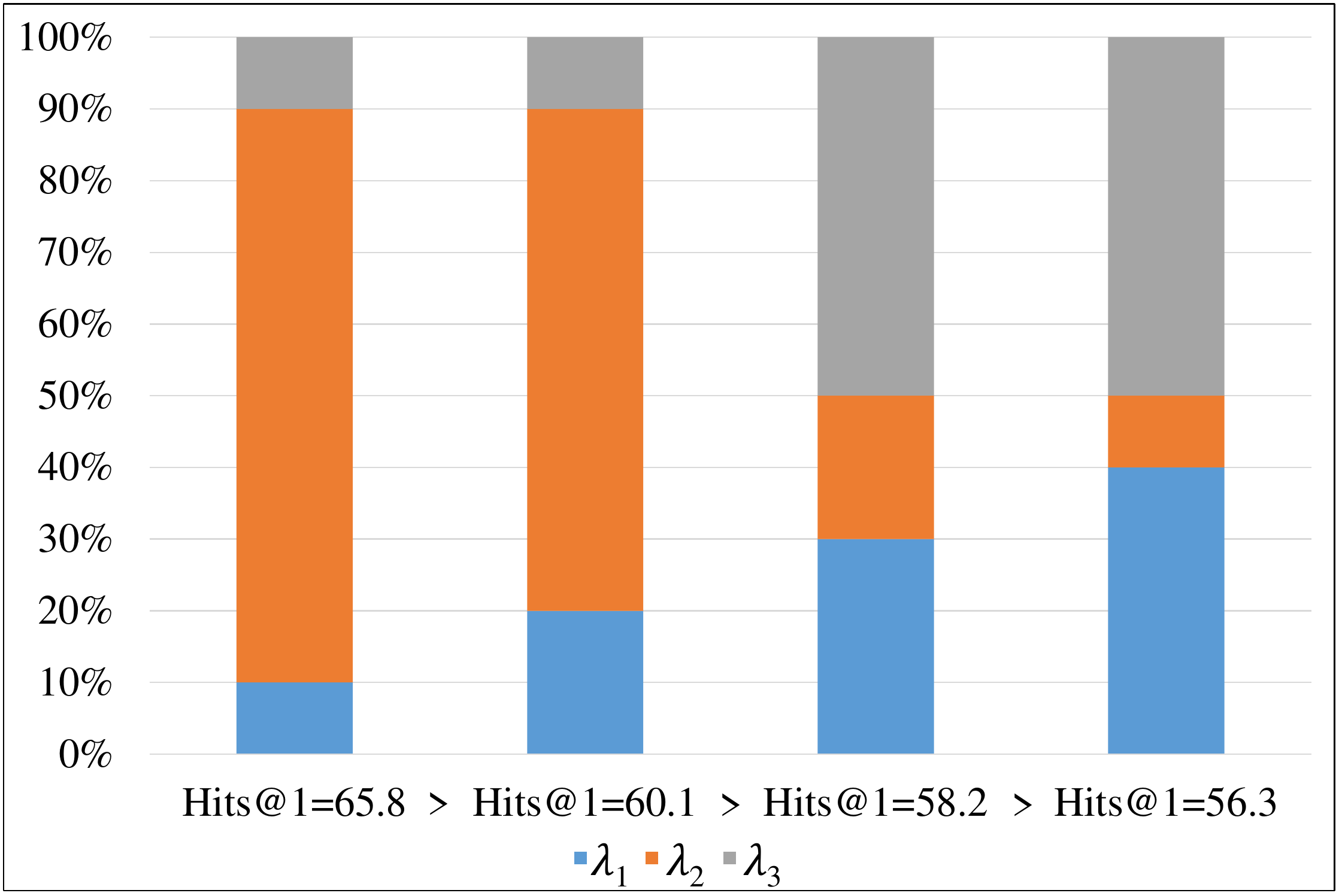}
	}
	\caption{Performance of MMKGR w.r.t. varied discount factor.}
	\label{fig:ablation}
\vspace{-0.5cm}
\end{figure}

In this study, we linearly combine destination reward, distance reward, and diverse  reward to form our 3D mechanism. The discount factor is equivalent to the combined weight of the reward.
Since the destination reward will return a relatively large reward value (i.e., 1) when it predicts the correct entity,  a smaller discount factor of destination reward is an appropriate choice. Otherwise, false positive target entities will be generated. 
Thus, the value of  $\lambda_{1}$ is set to some small values, i.e., 0.1, 0.2, 0.3, 0.4. 
Fig. 12 shows the highest Hits@1 in different number $\lambda$ combinations.
We can observe that the optimal  values of $\lambda_{1}$, $\lambda_{2}$, and $\lambda_{3}$ on different datasets are 0.1, 0.8, and 0.1 respectively. In addition, the increase of $\lambda_{1}$ will lead to the increase of $\lambda_{3}$ in a combination.
This is because high reward for reaching the target entity will make it easier to fall into the  local optimal path, and the diverse reward can solve the problem.

\section{Conclusion And Future Work}
In this work, we study the problem \textit{how to effectively leverage multi-modal auxiliary features to complete multi-hop KG reasoning}, which is an unexplored problem.
An effective model MMKGR is proposed, 
which outperforms the state-of-the-art approaches in the MKG reasoning task.
In MMKGR, we first perform feature extraction and multi-modal fine-grained fusion for structural data and multi-modal auxiliary data.
In addition, a unified gate-attention network is used to generate multi-modal complementary features. 
Next, these features are fed into a complementary feature-aware RL  framework to predict the missing elements. 
Extensive experiments demonstrate the rationality and effectiveness of MMKGR. In future work, we would like to construct some novel MKGs to study multi-hop reasoning. How to infer missing triplets over few-shot relations on MKGs, still awaits further exploration. 

\section{Acknowledgment}
This work was supported by the National Natural Science Foundation of China No. 61902270,  the Major Program of the Natural Science Foundation of Jiangsu Higher Education Institutions of China No. 19KJA610002, Australian Research Council Nos.FT210100624, DP190101985.

\bibliographystyle{IEEEtranS}
\bibliography{conference_101719}
\end{document}